\documentclass[dvipsnames]{article}

\usepackage[final]{openbmb_2025}

\usepackage{amsmath}
\usepackage{amssymb}
\usepackage{mathtools}
\usepackage{amsthm}
\usepackage{multirow}
\usepackage{makecell}
\usepackage{subcaption}
\usepackage{wrapfig}
\usepackage{graphicx}
\usepackage{pifont}

\usepackage[utf8]{inputenc} 
\usepackage[T1]{fontenc}    
\usepackage{url}            
\usepackage{booktabs}       
\usepackage{amsfonts}       
\usepackage{nicefrac}       
\usepackage{microtype}      

\usepackage{color, soul}
\usepackage[most,breakable]{tcolorbox}
\usepackage{xcolor}

\usepackage[labelfont=bf]{caption}
\usepackage{enumitem}
\usepackage{sectsty}

\usepackage{fancyhdr}
\renewcommand{\headrulewidth}{1pt}
\makeatletter
\def\headrule{{\if@fancyplain\let\headrulewidth\plainheadrulewidth\fi
\hrule\@height\headrulewidth\@width\textwidth \vskip-\headrulewidth}}
\makeatother

\definecolor{BMBDarkBlue}{HTML}{315EFE}
\definecolor{BMBLightBlue}{HTML}{00D3ED}
\usepackage[colorlinks, linkcolor=BMBDarkBlue, anchorcolor=BMBDarkBlue, citecolor=BMBDarkBlue, urlcolor=BMBDarkBlue]{hyperref}

\sectionfont{\color{BMBDarkBlue}\fontfamily{zi4}\selectfont}
\subsectionfont{\color{BMBDarkBlue}\fontfamily{zi4}\selectfont}
\subsubsectionfont{\color{BMBDarkBlue}\fontfamily{zi4}\selectfont}

\newtcolorbox{mytheorem}{
  colback=gray!5,       
  colframe=gray!80,     
  boxrule=0.5pt,        
  arc=4pt,              
  left=4pt,             
  right=4pt,            
  top=4pt,              
  bottom=4pt,           
}

\newcommand{\fancyheadname}{\textit{\textbf{Ultra-FineWeb}}}
\newcommand{\mydataset}[0]{Ultra-FineWeb}

\title{Ultra-FineWeb: Efficient Data Filtering and Verification for High-Quality LLM Training Data}

\author{%
Yudong Wang\textsuperscript{\rm 1}, 
Zixuan Fu\textsuperscript{\rm 2,3}, 
Jie Cai\textsuperscript{\rm 1},
Peijun Tang\textsuperscript{\rm 1},
Hongya Lyu\textsuperscript{\rm 1},
Yewei Fang\textsuperscript{\rm 1},\\
\textbf{
Zhi Zheng\textsuperscript{\rm 1},
Jie Zhou\textsuperscript{\rm 1},
Guoyang Zeng\textsuperscript{\rm 1},
Chaojun Xiao\textsuperscript{\rm 2},
Xu Han\textsuperscript{\rm 2},
Zhiyuan Liu\textsuperscript{\rm 2}}\\
\textsuperscript{\rm 1}ModelBest Inc.~~ \textsuperscript{\rm 2}Tsinghua University~~ \textsuperscript{\rm 3}Soochow University \\
\texttt{wangyudong@modelbest.cn}\quad \texttt{xcjthu@gmail.com}\\
\texttt{\{han-xu,liuzy\}@tsinghua.edu.cn}
}

\def\huggingface{\raisebox{-1.5pt}{\includegraphics[height=1.05em]{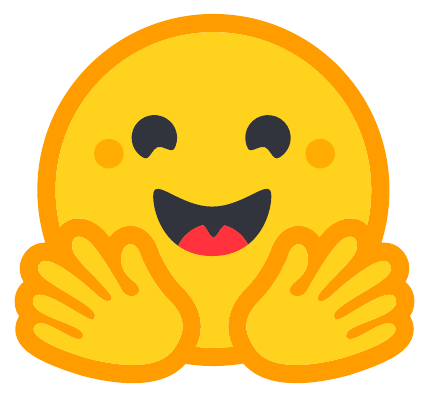}}}
\newcommand{\hflink}{https://huggingface.co/datasets/openbmb/UltraFineWeb}

\begin{document}

\maketitle
\thispagestyle{fancy} 

\vspace{1em}
\begin{abstract}
Data quality has become a key factor in enhancing model performance with the rapid development of large language models (LLMs). 
Model-driven data filtering has increasingly become a primary approach for acquiring high-quality data. 
However, it still faces two main challenges: (1) the lack of an efficient data verification strategy makes it difficult to provide timely feedback on data quality; and (2) the selection of seed data for training classifiers lacks clear criteria and relies heavily on human expertise, introducing a degree of subjectivity.
To address the first challenge, we introduce an efficient verification strategy that enables rapid evaluation of the impact of data on LLM training with minimal computational cost.
To tackle the second challenge, we build upon the assumption that high-quality seed data is beneficial for LLM training, and by integrating the proposed verification strategy, we optimize the selection of positive and negative samples and propose an efficient data filtering pipeline. 
This pipeline not only improves filtering efficiency, classifier quality, and robustness, but also significantly reduces experimental and inference costs.
In addition, to efficiently filter high-quality data, we employ a lightweight classifier based on \textit{fastText}, and successfully apply the filtering pipeline to two widely-used pre-training corpora, \textit{FineWeb} and \textit{Chinese FineWeb} datasets, resulting in the creation of the higher-quality \textit{\textbf{Ultra-FineWeb}} dataset. \mydataset{} contains approximately $1$ trillion English tokens and $120$ billion Chinese tokens.
Empirical results demonstrate that the LLMs trained on \mydataset{} exhibit significant performance improvements across multiple benchmark tasks, validating the effectiveness of our pipeline in enhancing both data quality and training efficiency.

\end{abstract}
\vspace{-1.5em}

\section{Introduction}
\vspace{-1.0em}
In recent years, Large Language Models (LLMs)~\citep{llama, han2021pre, instructGPT, minicpm, internlm, internlm2, qwen_tech_report} have achieved remarkable breakthroughs, demonstrating their powerful capabilities in various fields such as code generation~\citep{AlchemistCoder,deepseekcoder}, logical reasoning~\citep{deepseekr1, OREAL}, and scientific research~\citep{LLM4SR,ultramedical}.
Existing studies have indicated that large-scale high-quality (information-intensive) pre-training data is a key factor in driving the continuous improvement of LLMs' capabilities~\citep{fineweb,dclm,phi1,densing_law}.

To construct information-intensive pretraining corpora, current predominant approaches involve selective filtering of massive and noisy internet data sources~\citep{common_crawl}.
Early approaches usually rely on heuristic filtering using hand-crafted rules~\citep {c4,redpajama,rae2021scaling,ccnet} and deduplication~\citep {deduplicating}.
With increasing demands for better data quality, these heuristic approaches cannot identify complex content noise and lead to suboptimal LLM performance. Thus, model-driven data filtering, which employs a neural classifier to select high-quality content, has emerged as a better choice~\citep{phi1, deepseekmath}.
Notably, datasets such as FineWeb-edu~\citep{fineweb}, Chinese FineWeb-edu~\citep{chinese_fineweb}, CCI3-HQ~\citep{cci3}, and DCLM~\citep{dclm} have demonstrated the efficacy of this paradigm by incorporating model-based classifiers following preprocessing stages, achieving not only substantial improvements in dataset quality but also measurable enhancements in downstream LLM performance across various benchmark tasks.
Nevertheless, existing model-driven filtering methods still suffer from two main challenges: (1)~There is a lack of efficient validation to quickly verify the filtering results, typically requiring large-scale training to observe the effect, resulting in high costs and low efficiency. (2)~They heavily rely on manually-selected seed data, and the data for training classifiers often depends on human expertise, introducing significant subjectivity.

To address these challenges, we design an efficient data filtering pipeline based on an efficient verification strategy. This verification approach enables impartial seed data selection and facilitates iterative data filtering processes.
Specifically, in contrast to conventional approaches that verify data quality by training LLMs from scratch using candidate corpora, our proposed efficient verification strategy leverages a nearly-trained LLM as a foundation. We incorporate candidate corpora during the final training steps and utilize the resulting performance improvement as a metric for assessing data quality. This verification strategy significantly enhances evaluation efficiency while maintaining quality assessment accuracy.
Based on our efficient verification strategy, we can impartially select high-quality seed data for classifier training. 
Building upon the assumption that ``high-quality seed data is beneficial for LLM training'', we develop and optimize the strategy for selecting classifier training seeds and recipes, while carefully curating balanced sets of both positive and negative samples to ensure classifier quality and robustness.
In addition, to effectively reduce computational cost, we adopt a lightweight classifier based on fastText~\citep{fasttext}. Compared to LLM-based classifiers~\citep{fineweb}, the fastText-based classifier demonstrates superior inference efficiency, enabling both the filtering of higher-quality training data for LLMs and significantly accelerating the high-quality data filtration pipeline.
We apply the proposed data filtering pipeline to the FineWeb~\citep{fineweb} and Chinese FineWeb~\citep{chinese_fineweb} datasets (source data from Chinese FineWeb-edu-v2, which includes IndustryCorpus2~\citep{industryCorpus2}, MiChao~\citep{michao}, WuDao~\citep{wudao}, SkyPile~\citep{skypile}, WanJuan~\citep{wanjuan}, ChineseWebText~\citep{chinesewebtext}, TeleChat~\citep{telechat}, and CCI3~\citep{cci3}), resulting in the creation of higher-quality \textit{Ultra-FineWeb-en} and \textit{Ultra-FineWeb-zh} datasets, collectively referred to as \textit{\textbf{Ultra-FineWeb}}.
Experimental results show that LLMs trained on \textit{\textbf{Ultra-FineWeb}} perform excellently across multiple benchmark tasks, providing empirical validation for the effectiveness of our high-quality data filtering pipeline and its efficiency in reducing computational costs.

Our main contributions are as follows. The datasets and classifier will be released to facilitate the development of LLMs.
\begin{itemize}[leftmargin=*]
    \item \textbf{Efficient Verification Strategy:} We propose a computationally efficient verification strategy that enables rapid evaluation of the impact of data on LLM training performance with minimal computational cost, significantly improving the efficiency of high-quality data filtering experiments.
    \item \textbf{Large-Scale High-Quality Pre-training Datasets:} We design and implement an efficient high-quality data filtering pipeline, applied to the FineWeb and Chinese FineWeb datasets, resulting in the creation of higher-quality {\textit{Ultra-FineWeb-en}} and {\textit{Ultra-FineWeb-zh}} datasets, collectively referred to as \textbf{\textit{\mydataset{}}}. \textbf{\textit{\mydataset{}}} contains approximately $1$ trillion English tokens and $120$ billion Chinese tokens, and can facilitate high-quality LLM training.
    \item \textbf{Lightweight Classifier:} The \textit{\mydataset{} classifier }significantly reduces inference costs, achieving superior performance on extracted text from the same data source, thus validating the effectiveness of our proposed data filtering pipeline in enhancing data quality and training efficiency.
\end{itemize}

\vspace{-1.0em}
\section{Methodology}
\begin{figure*}[t]
    \centering
    \includegraphics[width=0.98\textwidth]{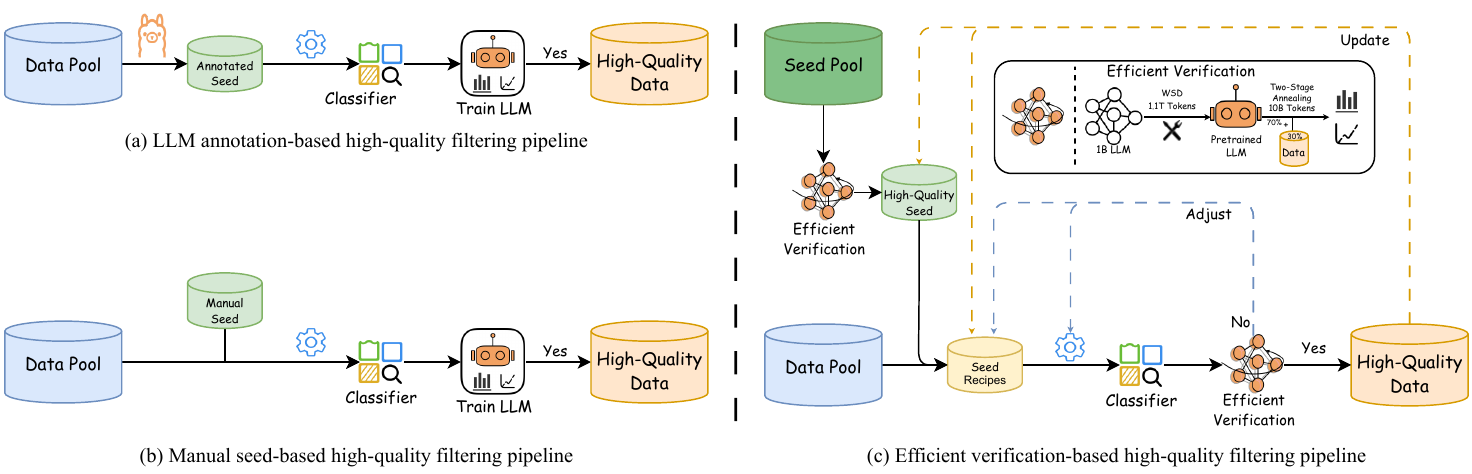}
    \caption{Comparison of High-Quality Data Filtering Pipelines. Traditional model-based data filtering methods (a) and (b) rely on human expertise for seed data selection and lack data quality verification.}
    \label{fig:overall}
\end{figure*}

\vspace{-1.0em}
This section introduces the design and implementation of our efficient, high-quality data filtering pipeline, with the overall workflow illustrated in Figure~\ref{fig:overall}(c). 
First, in Section \ref{method:efficient_verification_strategy}, we present an Efficient Verification Strategy that significantly reduces experimental costs while ensuring the reliability of evaluation results.
Subsequently, Section \ref{method:classifier_seed} outlines our methodology for selecting positive sample seed data for classifier training. 
Finally, Sections \ref{method:classifier_recipes} and \ref{method:fasttext_model} introduce classifier training recipes and fastText-based quality filtering, respectively, which together ensure optimal data selection quality and inference efficiency.

\vspace{-1.0em}
\subsection{Overall Workflow}
\vspace{-1.0em}
The overall workflow of the proposed efficient verification-based high-quality filtering pipeline is illustrated in Figure~\ref{fig:overall}(c). 
We begin by constructing an initial candidate seed pool and applying our efficient verification strategy to identify high-quality samples that significantly improve training performance. 
These verified samples serve as positive seeds for training a classifier, while negative samples are randomly selected from the raw data pool to create a balanced training set.
During the classifier filtering stage, we sample a small subset from the raw data pool and validate the classifier's selections using our efficient verification strategy to assess its effectiveness. 
Based on verification results, we iteratively update the high-quality seed pool, adjust the ratio of positive and negative samples, and fine-tune classifier training hyperparameters to optimize the data selection strategy.
Only classifiers demonstrating stable and reliable performance in efficient verification are deployed for full-scale data selection and subsequent model training, thereby significantly reducing computational costs while maintaining high data quality.

\vspace{-1.0em}
\subsection{Efficient Verification Strategy}
\vspace{-1.0em}
\label{method:efficient_verification_strategy}
Validating the effectiveness of training data typically requires significant computational resources. For instance, training a 1 billion (B) LLM on 100B tokens requires approximately 1,200 H100 GPU hours (equivalent to 64 GPUs running continuously for nearly 19 hours).
This computational burden becomes particularly prohibitive when iteratively developing high-quality data classifiers. 
Moreover, large-scale training validation proves impractical for smaller datasets, as models trained with limited token counts fail to exhibit statistically significant performance differences, with training instability further compromising result reliability.
This limitation is evident in our comparative analysis of FineWeb and FineWeb-edu~\citep{fineweb}. When trained from scratch with 8 billion tokens, FineWeb-edu achieves superior performance on HellaSwag~\citep{hellaswag}, while at 380 billion tokens, FineWeb demonstrates better results across multiple benchmarks, including Winogrande~\citep{winogrande}, HellaSwag~\citep{hellaswag}, and PIQA~\citep{piqa}, highlighting the inconsistency in evaluation outcomes based on training scale\footnote{\url{https://huggingface.co/spaces/HuggingFaceFW/blogpost-fineweb-v1}}.

Inspired by Llama 3.1~\citep{llama3}, we design an Efficient Verification Strategy. We begin by training a 1B LLM on 1.1 trillion (T) tokens using a WSD scheduler~\citep{minicpm} (comprising stable training on 1T tokens, followed by decay training on 0.1T tokens).
Based on this pretrained LLM, we then implement a two-stage annealing process with 10B tokens, allocating 30\% of the weight to the verification data, while keeping the remaining 70\% for the default mixed data ratio. Model details and training hyperparameters can be found in Appendix \ref{appendix:implement_details}. 
As shown in Table~\ref{tab:comparison_train_cost}, this optimized strategy reduces computational costs from 1,200 to approximately 110 H100 GPU hours (equivalent to less than 3.5 hours on 32 GPUs), significantly reducing training costs and effectively improving the efficiency and iterability of the filtering process, with the two-stage annealing results using the original mixed data ratio as the baseline.
This strategy allows for efficient assessment of the impact of verification data across various evaluation dimensions. 
To validate the reliability of this strategy, we compare the results of training 100B tokens from scratch on the 1B LLM using FineWeb and FineWeb-edu, respectively. As shown in Table \ref{tab:appendix_efficient_verification_result}, the results follow similar trends, with further experimental analysis provided in Appendix \ref{appendix:efficient_verification_result}.

\begin{table}[!htb]
    \centering
    \small
    \caption{Comparison of computational costs across different verification strategies on a 1B LLM.}
    \renewcommand{\arraystretch}{1.2}
    \begin{tabular}{lccc}
        \toprule
         & 100B from scratch &	380B from scratch &	Efficient Verification Stratege  \\
        \midrule
        GPU Hours &  1,200	        &  4,600	&  \textbf{110} \\
        \bottomrule
    \end{tabular}
    \label{tab:comparison_train_cost}
\end{table}

\vspace{-1.0em}
\subsection{Classifier Training Seeds}
\vspace{-1.0em}
\label{method:classifier_seed}
The effectiveness of high-quality data classifiers fundamentally depends on the selection of superior positive training samples. 
As illustrated in Figure~\ref{fig:overall}(a), datasets such as FineWeb-edu~\citep{fineweb}, Chinese-FineWeb-edu~\citep{chinese_fineweb}, and CCI3-HQ~\citep{cci3} employ LLM annotation-based frameworks to partially label source-consistent data, generating ``seed data''. 
In contrast, Figure~\ref{fig:overall}(b) demonstrates manual seed-based filtering (DCLM's~\citep{dclm}) pipeline, which relies on manual curation for positive sample selection, focusing specifically on instruction-formatted data by incorporating samples from OpenHermes 2.5 (OH-2.5)~\citep{OH25} and high-quality posts from the \texttt{r/ExplainLikeImFive} (ELI5) subreddit.

Although both pipelines demonstrate distinct advantages in selecting positive samples, they are accompanied by inherent limitations.
The LLM annotation-based pipeline can effectively filter high-quality samples from source-consistent data, but its performance is constrained by the scoring criteria of the LLM, potentially introducing systematic biases and annotation noise. 
Furthermore, classifiers trained exclusively on source-consistent data often exhibit limited generalization capabilities and poor robustness. 
Conversely, manual curation faces significant methodological challenges: the effectiveness of seed data is difficult to assess before classifier training, and its validation relies heavily on the performance of LLMs trained on the filtered data. These constraints lead to high computational costs and reduced adaptability across different tasks.

Based on these considerations, we propose a key assumption: high-quality seed data that enhances LLM performance will yield classifiers capable of identifying similarly beneficial training data.
As illustrated in Figure~\ref{fig:overall} (c), we implement our Efficient Verification Strategy to rapidly evaluate and validate seed data quality within the candidate pool, ensuring the selection of samples that can improve LLM training results. 
This pipeline not only ensures superior data quality but also optimizes filtration efficiency, thereby generating more reliable positive samples for classifier training. 
Furthermore, to enhance classifier robustness, we expand negative sample selection beyond source-consistent data. Experimental results further demonstrate that incorporating diverse data sources for negative samples can improve the generalizability of the classifier.

\vspace{-1.0em}
\subsection{Classifier Training Recipes}
\vspace{-1.0em}
\label{method:classifier_recipes}
We evaluate a large pool of candidate seed data and ultimately select those with clear effectiveness as positive samples. The positive samples include: (1) LLM-annotated data with scores above 4\footnote{\url{https://huggingface.co/datasets/HuggingFaceFW/fineweb-edu-llama3-annotations}}\textsuperscript{,}\footnote{\url{https://huggingface.co/datasets/BAAI/CCI3-HQ-Annotation-Benchmark}}; (2) instruction-formatted datasets such as OH-2.5 and ELI5; (3) authentic textbook data; (4) LLM-synthesized educational data; and (5) high-quality web content obtained through targeted crawling.
For negative samples, we incorporate raw data from diverse sources, including English corpora (FineWeb~\citep{fineweb}, C4~\citep{c4}, Dolma~\citep{dolma}, Pile~\citep{pile}, and RedPajama~\citep{redpajama}) and Chinese datasets (IndustryCorpus2~\citep{industryCorpus2}, MiChao~\citep{michao}, WuDao~\citep{wudao}, SkyPile~\citep{skypile}, WanJuan~\citep{wanjuan}, ChineseWebText~\citep{chinesewebtext}, TeleChat~\citep{telechat}, and CCI3~\citep{cci3}) in the initial iteration. 
To maintain dataset diversity and balance, we implement a uniform distribution strategy, with underrepresented categories undergoing 3-5 rounds of strategic resampling.

Subsequently, we conduct a single iteration of the classifier, utilizing its current predictions as training data for the next round. 
However, empirical results indicate that the iterative process only contributed meaningfully in the first round, as subsequent updates do not yield further performance improvements and, in some cases, even lead to a decrease in LLM performance. 
Our analysis shows that classifier improvement primarily depends on the seed data selection, rather than iterative refinement using inferred samples. 
Interestingly, we find that intersecting high-quality data filtered by multiple classifiers consistently improves LLM performance.

\vspace{-1.0em}
\subsection{FastText-based Quality Filtering}
\vspace{-1.0em}
\label{method:fasttext_model}
Current high-quality data classifiers are primarily divided into LLM-based~\citep{fineweb, chinese_fineweb, cci3} and fastText-based~\citep{dclm, deepseekmath, deepseekcoder} methods. 
While LLM-based classifiers are effective, they need significantly higher inference costs. 
To address this, we adopt a fastText-based classifier, which significantly reduces inference costs while maintaining competitive performance under certain conditions. 
This approach not only minimizes resource consumption but also speeds up data filtering experiments.
For instance, as shown in Table~\ref{tab:comparison_infer_cost}, processing 15T tokens with an LLM-based classifier requires approximately 6,000 H100 GPU hours, while fastText can complete the same task on a non-GPU machine with just 80 CPUs in 1,000 hours, significantly improving efficiency.
Notably, most of our large-scale experiments are conducted in a distributed manner using a Spark\footnote{\url{https://spark.apache.org/}} cluster.

\begin{table}[!htb]
    \centering
    \small
    \caption{Comparison of inference costs for different model-based classifiers on 15T tokens}
    \renewcommand{\arraystretch}{1.2}
    \begin{tabular}{lcc}
        \toprule
        & LLM-based Classifier & fastText-based Classifier \\
        \midrule
        GPU Used               & \ding{51}                 & \ding{55} \\
        CPU Used               & \ding{51}                 & \ding{51} \\
        Processing Time (Hours)& 6,000                     & \textbf{1,000}     \\
        \bottomrule
    \end{tabular}
    \label{tab:comparison_infer_cost}
\end{table}

For data preprocessing, we implement several key steps, including removing redundant empty lines and extra spaces, stripping diacritics, and converting all English text to lowercase. 
Additionally, we adopt the DeepSeek-V2 tokenizer~\citep{deepseekv2}, which outperforms traditional tokenization methods (such as space-based tokenization for English and Jieba\footnote{\url{https://pypi.org/project/jieba/}} for Chinese). 
Meanwhile, we preserve structural information such as \texttt{\textbackslash n}, \texttt{\textbackslash t}, and \texttt{\textbackslash r}. 
To ensure dataset integrity and balance, the final training set comprised 600K samples, evenly split between positive and negative examples.

For training details, we trained a fastText classifier with a vector dimension of 256, a learning rate of 0.1, a maximum word n-gram length of 3, a minimum word occurrence threshold of 5, and a total of 3 training epochs.
Additionally, during inference, we maintain the default threshold of 0.5 to simplify operations and ensure experimental consistency, avoiding the need for additional tuning steps.

\vspace{-1.0em}
\section{Experiments}
\vspace{-1.0em}
In this section, we first detail the experimental settings in Section~\ref{exp:exp_set}, including the training configuration, data composition, and evaluation metrics.
Then, in Section~\ref{exp:overall_result}, we present the overall experimental results, highlighting the performance comparisons between individual datasets and mixed datasets.
These results demonstrate that \textit{\mydataset{}}, obtained through our efficient data filtering pipeline, exhibits superior quality to other datasets derived from the same data source, with the corresponding trained models achieving enhanced performance. 
Finally, in Section~\ref{exp:ab_study}, we perform extensive ablation studies to further evaluate the effectiveness of \textit{\mydataset{}}, examining the impact of seed and recipe selection strategies for classifier training and the quality of the intersection of positive samples filtered by multiple classifiers.

\vspace{-1.0em}
\subsection{Experimental Setting}
\vspace{-1.0em}
\label{exp:exp_set}

\textbf{Model Training Configuration.} In our experiments, all models are trained using the open-source Megatron-LM library~\citep{megatron}. We utilize the MiniCPM-1.2B model architecture with the MiniCPM3-4B tokenizer. 
Each experiment involves training on 100B tokens (though the actual number is 104B tokens, calculated as $4096 \times 1024 \times 26000 = 104$B tokens; for simplicity, we refer to it as $100$B), allowing for comprehensive data performance validation within computationally efficient parameters.
Key training parameters include a sequence length of 4096, weight decay of 0.1, and a gradient clipping threshold of 1.0. We employ a global batch size of 1,024 across 26,000 training steps. 
The learning rate follows a cosine decay schedule, with a warm-up phase of 1,000 steps. The initial learning rate is set to 1e-5, the maximum learning rate to 1e-2, and the final learning rate to 1e-3. 
To enhance training stability, we use Maximal Update Parameterization (MuP)~\citep{mup}. 
Additionally, we save a checkpoint every 1,000 steps (approximately 4B tokens) for analysis during the training process. 
Detailed model configurations are provided in Table \ref{tab:model_cfg}, where $Params.$, $Vocab.$, $d_m$, $d_{ff}$, $d_h$, $n_{head}$, $n_{kv}$, and $n_{Layer}$ represent the total number of non-embedding parameters, vocabulary size, model hidden dimension, feedforward layer bottleneck dimension, attention head dimension, number of queries, number of key/values, and the number of layers, respectively.

\begin{table}[!h]
    \centering
    \small
    \caption{Model Configurations for the MiniCPM-1.2B model.}
    \renewcommand{\arraystretch}{1.2}
    \begin{tabular}{ccccccccc}
        \toprule
        Name & $Params.$ & $Vocab.$ & $d_m$ & $d_{ff}$ & $d_h$ & $n_{head}$ & $n_{kv}$ & $n_{Layer}$ \\
        \midrule
        MiniCPM-1.2B & 1,247,442,432 & 73448 & 1,536 & 3,840 & 64 & 24 & 8 & 52 \\
        \bottomrule
    \end{tabular}
    \label{tab:model_cfg}
\end{table}

\textbf{Dataset Composition.} We conduct two types of experiments for evaluating the datasets generated by our pipeline:
\begin{itemize}[leftmargin=*]
    \item \textbf{Individual Data Experiments:} We perform isolated training runs using single datasets, facilitating direct comparisons between differently processed data from identical sources. For English datasets, FineWeb is chosen as the source dataset, and comparisons are made with FineWeb-edu and Ultra-FineWeb-en. For Chinese datasets, Chinese FineWeb is selected with comparisons to Chinese FineWeb-edu-v2 and Ultra-FineWeb-zh. In the ablation studies, we primarily use individual data experiments for analysis.
    \item \textbf{Mixed Data Experiments:} Similar to the CCI3-HQ~\citep{cci3} experiment, we use a mix of 60\% English data, 30\% Chinese data, and 10\% code data. The English-Chinese comparisons involve three dataset combinations: (1) FineWeb and Chinese FineWeb, (2) FineWeb-edu and Chinese FineWeb-edu-v2, and (3) Ultra-FineWeb-en and Ultra-FineWeb-zh. The code data is sourced exclusively from the StarCoder-v2 dataset~\citep{starcoder2}, maintaining consistent proportions across all experimental conditions.
\end{itemize}

\textbf{Evaluation Metrics.} We employ the Lighteval~\citep{lighteval} library for model evaluation, mirroring the setup used with FineWeb~\citep{fineweb} and CCI3-HQ~\citep{cci3}. All evaluation metrics are based on a zero-shot setting. The evaluation metrics include:
\begin{itemize}[leftmargin=*]
    \item \textit{Average$_{English}$}: Average score across standard English metrics including MMLU~\citep{mmlu}, ARC-C~\citep{arc}, ARC-E~\citep{arc}, CommonSenseQA~\citep{commonsenseqa}, HellaSwag~\citep{hellaswag}, OpenbookQA~\citep{openbookqa}, PIQA~\citep{piqa}, SIQA~\citep{siqa}, and Winogrande~\citep{winogrande}.
    \item \textit{Average$_{Chinese}$}: Average score of Chinese metrics, including C-Eval~\citep{ceval} and CMMLU~\citep{cmmlu}.
    \item \textit{Average}: The combined average score of all the above evaluation metrics.
\end{itemize}

\vspace{-1.0em}
\subsection{Overall Results}
\vspace{-1.0em}
\label{exp:overall_result}

\begin{figure}[!htb]
    \centering
    \begin{subfigure}{0.48\textwidth}
        \includegraphics[width=\linewidth]{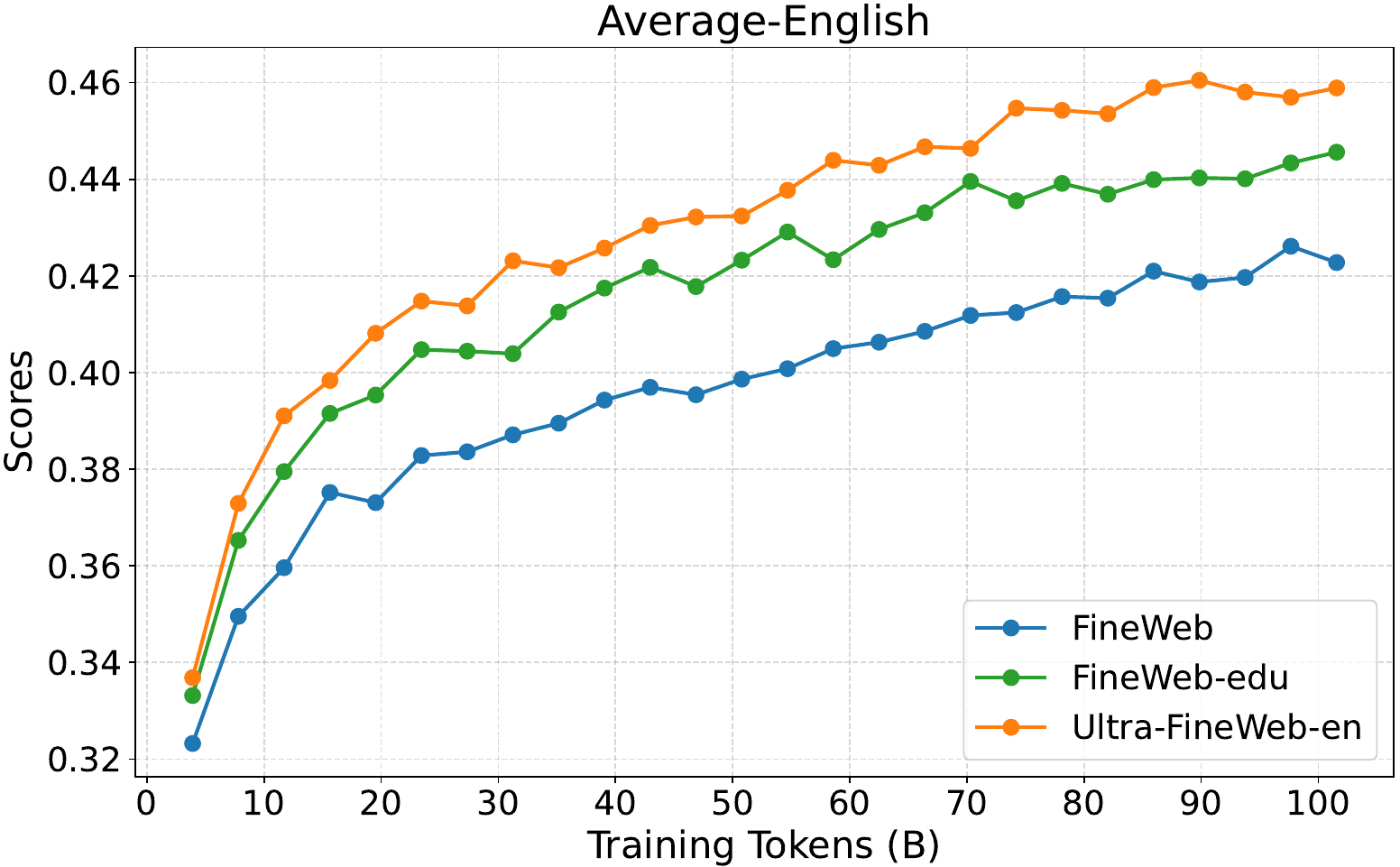}
        \captionsetup{justification=centerlast, singlelinecheck=false, width=\linewidth}
        \caption{Results on English datasets.}
    \end{subfigure}
    \begin{subfigure}{0.48\textwidth}
        \includegraphics[width=\linewidth]{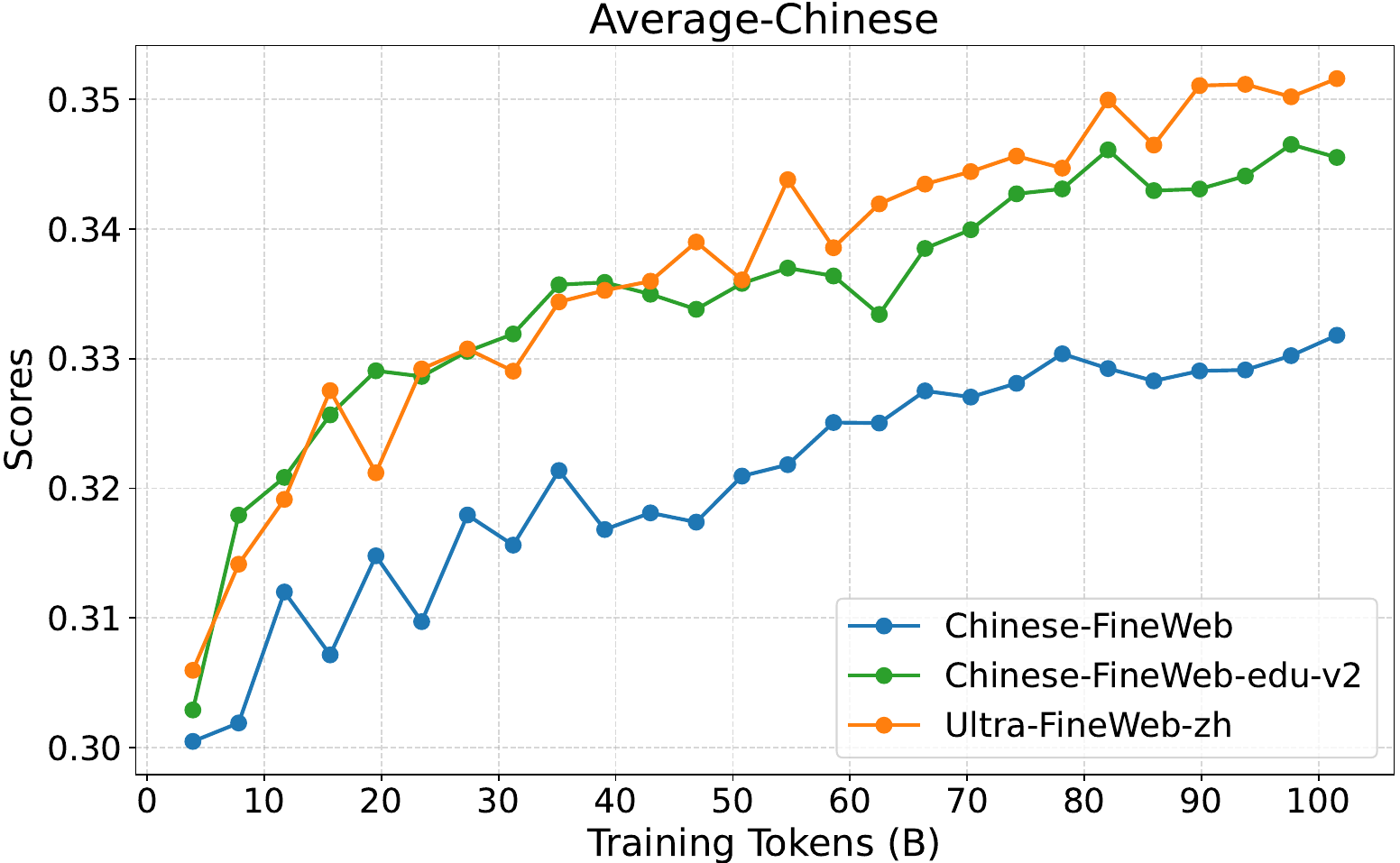}
        \captionsetup{justification=centerlast, singlelinecheck=false, width=\linewidth}
        \caption{Results on Chinese datasets.}
    \end{subfigure}
    \caption{Average scores at each checkpoint for different individual datasets.}
    \label{fig:single_result}
\end{figure}

\textbf{Individual Dataset Results.} We compare the performance of models trained on 100B tokens using data extracted from the FineWeb and Chinese FineWeb sources, using three different approaches: raw data, LLM-based classifiers (-edu), and the fastText-based classifier trained via the Efficient Data Filtering Pipeline (Ultra-). 
As shown in Tables~\ref{tab:en_result} and~\ref{tab:zh_result}, on the English Metrics, Ultra-FineWeb-en demonstrates significant improvements in performance on multiple tasks, including MMLU, ARC-C, ARC-E, CommonSenseQA, and OpenBookQA. 
Specifically, Ultra-FineWeb outperforms FineWeb in these tasks, with only a slight drop of 0.15 percentage points ($pp$) in HellaSwag compared to FineWeb, but a 0.6$pp$ improvement over FineWeb-edu. 
The English average score for Ultra-FineWeb-en (45.891$pp$) is 3.61$pp$ higher than that of FineWeb (42.287$pp$) and 1.3$pp$ higher than FineWeb-edu (44.560$pp$). 
On the Chinese metrics, Ultra-FineWeb-zh also outperforms both FineWeb-zh and FineWeb-edu-zh on C-Eval and CMMLU. 
Specifically, Ultra-FineWeb-zh improves by 0.31$pp$ and 3.65$pp$ over Chinese FineWeb and Chinese FineWeb-edu-v2 on C-Eval and CMMLU, respectively, and by 0.09$pp$ and 0.13$pp$ compared to FineWeb-edu-zh. 
The Chinese average score for Ultra-FineWeb-zh increases by 1.98$pp$ and 0.61$pp$, respectively, compared to FineWeb-zh and FineWeb-edu-zh. These results indicate that our proposed High-Quality Data Filtering Pipeline significantly improves data quality, leading to notable improvements in model performance. 
Additionally, we evaluate the performance at each training checkpoint. As shown in Figure~\ref{fig:single_result}, Ultra-FineWeb-en surpasses both FineWeb and FineWeb-edu early in the training process, while Ultra-FineWeb-zh demonstrates a marked improvement in Chinese average scores after 40B tokens of training.

\begin{table*}[t]
    \centering
    \small
    \caption{Comparison of individual results on English datasets.}
    \renewcommand{\arraystretch}{1.2}
    \begin{tabular}{llll}
        \toprule
        Metrics                         & FineWeb           & FineWeb-edu                           & Ultra-FineWeb-en \\
        \midrule
        MMLU        	                &  28.84	        &  31.80$_{\textcolor{red}{+2.96}}$	    &  \textbf{32.24}$_{\textcolor{red}{+3.4}}$    \\
        ARC-C	                        &  25.17	        &  34.56$_{\textcolor{red}{+9.39}}$	    &  \textbf{35.67}$_{\textcolor{red}{+10.5}}$   \\
        ARC-E	                        &  59.18	        &  69.95$_{\textcolor{red}{+10.77}}$    &  \textbf{70.62}$_{\textcolor{red}{+11.44}}$  \\
        CommonSenseQA	                &  34.32	        &  31.53$_{\textcolor{blue}{-2.79}}$    &  \textbf{36.45}$_{\textcolor{red}{+2.13}}$   \\
        HellaSwag	                    &  \textbf{42.91}   &  42.17$_{\textcolor{blue}{-0.74}}$    &  42.76$_{\textcolor{blue}{-0.15}}$  \\
        OpenbookQA	                    &  22.20	        &  25.20$_{\textcolor{red}{+3.00}}$	    &  \textbf{26.20}$_{\textcolor{red}{+4.00}}$   \\
        PIQA	                        &  73.29	        &  72.14$_{\textcolor{blue}{-1.15}}$	&  \textbf{73.67}$_{\textcolor{red}{+0.38}}$   \\
        SIQA	                        &  38.95	        &  38.13$_{\textcolor{blue}{-0.82}}$	&  \textbf{39.61}$_{\textcolor{red}{+0.66}}$   \\
        Winogrande	                    &  55.64	        &  55.56$_{\textcolor{blue}{-0.08}}$	&  \textbf{55.80}$_{\textcolor{red}{+0.16}}$   \\
        \midrule
        \textit{Average$_{English}$}    &  42.278	        &  44.560$_{\textcolor{red}{+2.282}}$	&  \textbf{45.891}$_{\textcolor{red}{+3.613}}$ \\
        \bottomrule
    \end{tabular}
    \label{tab:en_result}
\end{table*}

\begin{table}[!htb]
    \centering
    \small
    \caption{Comparison of individual results on Chinese datasets.}
    \renewcommand{\arraystretch}{1.2}
    \begin{tabular}{llll}
        \toprule
        Metrics	                            &  Chinese-FineWeb	& Chinese-FineWeb-edu-v2 	          & Ultra-FineWeb-zh                   \\
        \midrule
        C-Eval	                            &  33.95	        & 34.17$_{\textcolor{red}{+0.22}}$	  & \textbf{34.26}$_{\textcolor{red}{+0.31}}$     \\
        CMMLU	                            &  32.41	        & 34.93$_{\textcolor{red}{+2.52}}$     & \textbf{36.06}$_{\textcolor{red}{+3.65}}$     \\
        \midrule
        \textit{Average$_{Chinese}$}        &  33.18	        & 34.55$_{\textcolor{red}{+1.370}}$    & \textbf{35.16}$_{\textcolor{red}{+1.980}}$     \\
        \bottomrule
    \end{tabular}
    \label{tab:zh_result}
\end{table}

\begin{figure}[htbp]
    \centering
    \includegraphics[width=1.0\textwidth]{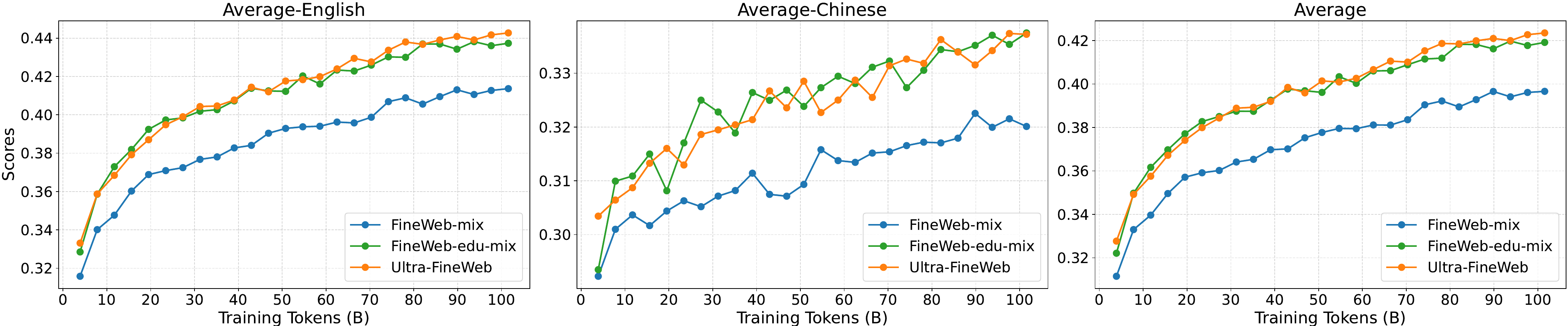}
    \caption{Average scores at each checkpoint for different mixed datasets.}
    \label{fig:mix_result}
\end{figure}

\textbf{Mixed Dataset Results.}
In the mixed data experiments, we compare the model performance on different evaluation sets after training 100B tokens with the original data, LLM-based classifier-extracted edu data, and our Ultra-FineWeb dataset, using the same training configuration. 
As shown in Table~\ref{tab:mix_result}, Ultra-FineWeb demonstrates significant performance improvements on multiple benchmarks. The average English score is 2.905$pp$ higher than FineWeb$_{mix}$ (41.366$pp$) and 0.538$pp$ higher than FineWeb-edu$_{mix}$ (43.733$pp$). 
For the Chinese evaluation set, Ultra-FineWeb achieves a 1.715$pp$ advantage over than FineWeb$_{mix}$ (32.01$pp$), while showing a marginal 0.025$pp$ decrease compared to FineWeb-edu$_{mix}$ (33.75$pp$). 
This minor discrepancy may stem from dataset weight setting or inherent training instability, warranting further investigation in future studies. 
The comprehensive analysis reveals Ultra-FineWeb's superior performance over both baseline and LLM-filtered datasets, demonstrating significant overall score improvements. 
Despite task-specific fluctuations, Ultra-FineWeb, generated through our Efficient Data Filtering Pipeline, consistently delivers effective performance enhancements. 
The line charts of checkpoint evaluations are shown in Figure~\ref{fig:mix_result}. 
In the early training phases, Ultra-FineWeb and FineWeb-edu$_{mix}$ exhibit comparable performance, but both outperform FineWeb$_{mix}$. Notably, Ultra-FineWeb starts to surpass FineWeb-edu$_{mix}$ after training approximately 60B tokens. 
As for Chinese evaluation metrics, both Ultra-FineWeb and FineWeb-edu$_{mix}$ demonstrate training fluctuations while maintaining substantial advantages over FineWeb$_{mix}$ throughout the training process.

\begin{table}[h]
    \centering
    \small
    \caption{Comparison of results on mixed datasets.}
    \renewcommand{\arraystretch}{1.2}
    \begin{tabular}{llll}
        \toprule
        Metrics                             & FineWeb$_{mix}$           & FineWeb-edu$_{mix}$                                   & Ultra-FineWeb \\
        \midrule
        MMLU	                            &  28.50	            &  \textbf{30.95}$_{\textcolor{red}{+2.45}}$	    &  30.94$_{\textcolor{red}{+2.44}}$                 \\
        ARC-C	                            &  24.15	            &  32.34$_{\textcolor{red}{+8.19}}$	                &  \textbf{33.36}$_{\textcolor{red}{+9.21}}$        \\
        ARC-E	                            &  55.60	            &  67.13$_{\textcolor{red}{+11.53}}$	            &  \textbf{67.97}$_{\textcolor{red}{+12.37}}$       \\
        CommonSenseQA	                    &  36.20	            &  35.79$_{\textcolor{blue}{-0.41}}$	            &  \textbf{37.18}$_{\textcolor{red}{+0.98}}$        \\
        HellaSwag	                        &  \textbf{40.28}	    &  40.21$_{\textcolor{blue}{-0.07}}$	            &  39.65$_{\textcolor{blue}{-0.63}}$                \\
        OpenbookQA	                        &  21.60	            &  23.80$_{\textcolor{red}{+2.20}}$	                &  \textbf{24.40}$_{\textcolor{red}{+2.80}}$        \\
        PIQA	                            &  71.11	            &  \textbf{71.22}$_{\textcolor{red}{+0.11}}$	    &  70.08$_{\textcolor{blue}{-1.03}}$                \\
        SIQA	                            &  39.76	            &  39.20$_{\textcolor{blue}{-0.56}}$	            &  \textbf{40.48}$_{\textcolor{red}{+0.72}}$        \\
        Winogrande	                        &  \textbf{55.09}	    &  52.96$_{\textcolor{blue}{-2.13}}$	            &  54.38$_{\textcolor{blue}{-0.71}}$                \\
        \midrule
        C-Eval                              &  33.79	            &  \textbf{34.32}$_{\textcolor{red}{+0.53}}$	    &  34.10$_{\textcolor{red}{+0.31}}$                 \\
        CMMLU                               &  30.23	            &  33.18$_{\textcolor{red}{+2.95}}$	                &  \textbf{33.35}$_{\textcolor{red}{+3.12}}$        \\
        \midrule
        \textit{Average$_{English}$}	    & 41.366	            &  43.733$_{\textcolor{red}{+2.367}}$	            &  \textbf{44.271}$_{\textcolor{red}{+2.905}}$      \\
        \textit{Average$_{Chinese}$}	    & 32.010	            &  \textbf{33.750}$_{\textcolor{red}{+1.740}}$	    &  33.725$_{\textcolor{red}{+1.715}}$               \\
        \textit{Average}                    & 39.665	            &  41.918$_{\textcolor{red}{+2.253}}$	           & \textbf{42.354}$_{\textcolor{red}{+2.689}}$        \\
        \bottomrule
    \end{tabular}
    \label{tab:mix_result}
\end{table}

\vspace{-1.0em}
\subsection{Ablation Study and Analysis}
\label{exp:ab_study}
\vspace{-1.0em}

\textbf{Analysis of Token Length Distributions. }
We first analyze the token length distributions across different datasets, as shown in Figure 4. 
For the English datasets, the token length distributions of Ultra-FineWeb-en and FineWeb are quite similar, while FineWeb-edu exhibits a rightward shift, indicating that the classifier tends to extract longer tokens. 
In terms of average token length, FineWeb has the shortest average, followed by Ultra-FineWeb, with FineWeb-edu having the longest average token length.
For the Chinese datasets, Ultra-FineWeb-zh and Chinese FineWeb exhibit similar token length distributions, while Chinese FineWeb-edu-v2 also shows a rightward shift. 
The average token length follows the order: Chinese FineWeb < Chinese FineWeb-edu-v2 < Ultra-FineWeb-zh. 
We believe these differences may stem from the inherent preference of LLM-based models, which tend to favor longer tokens in their scoring. 
Additionally, this phenomenon might be further influenced by training recipes, as LLM-based models label data from the same source, typically assigning lower scores to shorter texts and higher scores to longer ones, leading classifiers to favor longer texts. 
In contrast, our data seeds are more diverse, making the classifiers less focused on token length, which results in the token length distribution of our extracted data aligning more closely with the original source data.

\begin{figure}[htbp]
    \centering
    \begin{subfigure}{0.48\textwidth}
        \includegraphics[width=\linewidth]{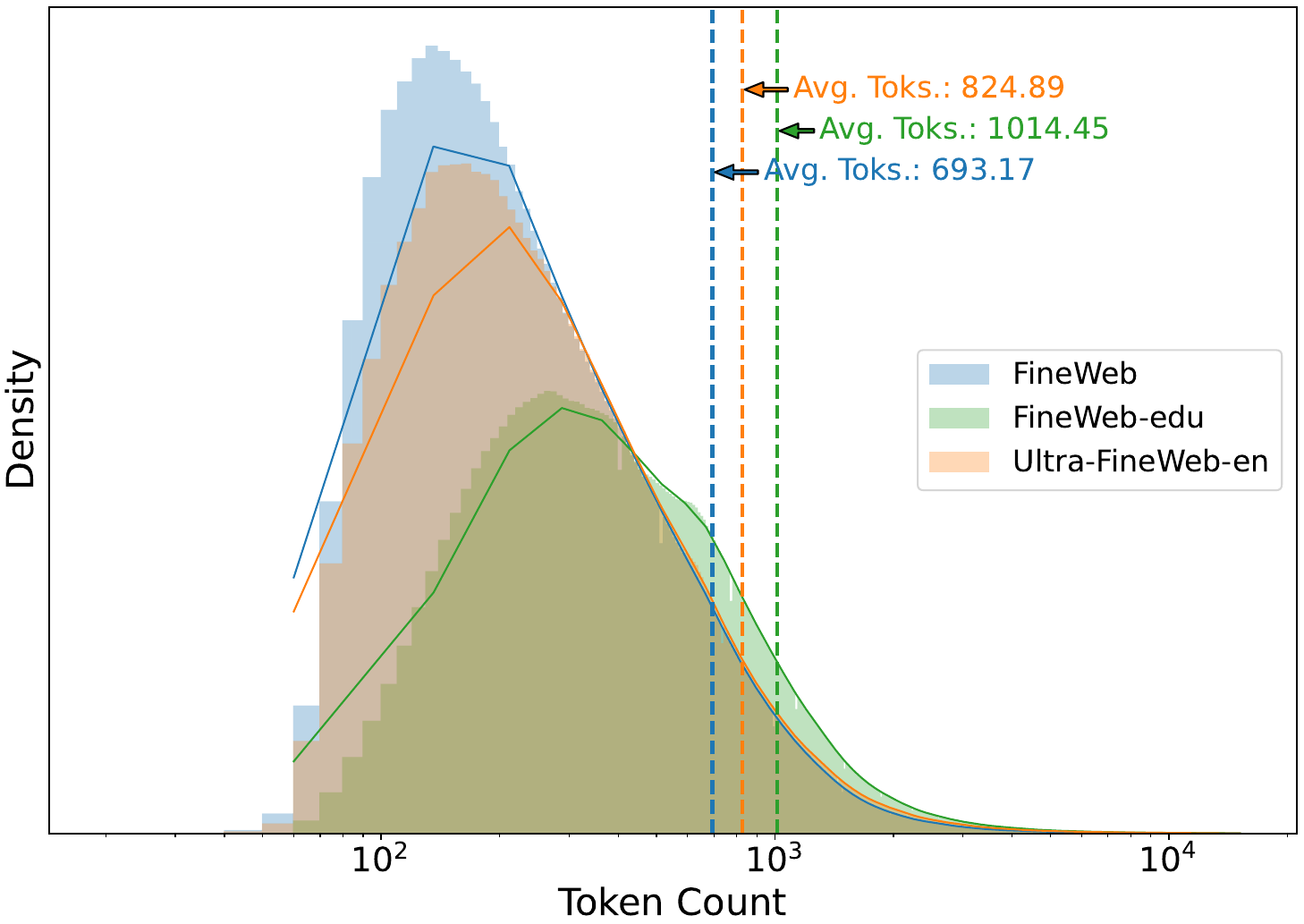}
        \captionsetup{justification=centerlast, singlelinecheck=false, width=\linewidth}
        \caption{Comparison of token length distributions on English datasets.}
    \end{subfigure}
    \begin{subfigure}{0.48\textwidth}
        \includegraphics[width=\linewidth]{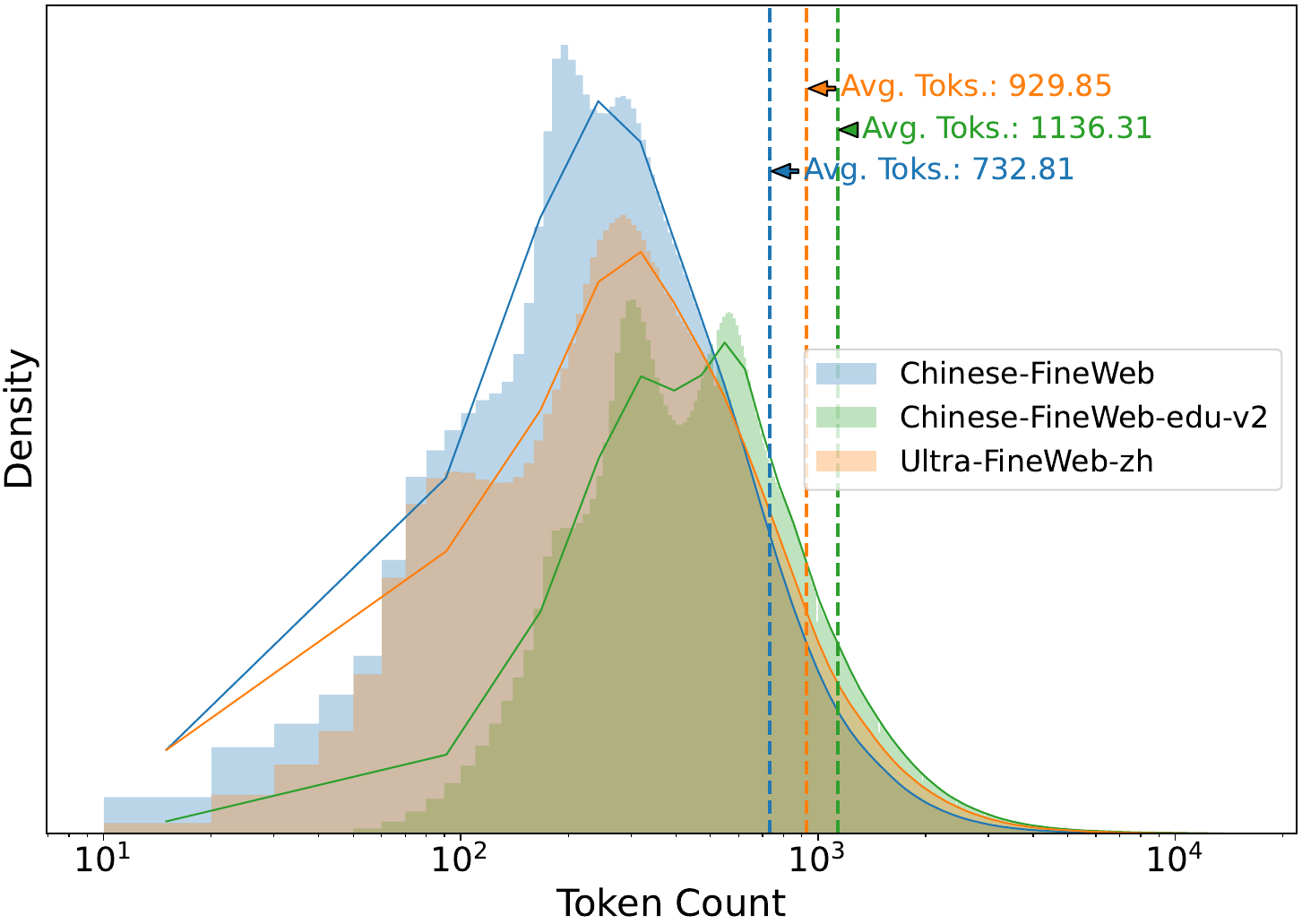}
        \captionsetup{justification=centerlast, singlelinecheck=false, width=\linewidth}
        \caption{Comparison of token length distributions on Chinese datasets.}
    \end{subfigure}
    \caption{Comparison of token length distributions across different datasets.}
    \label{fig:token_ann}
\end{figure}

\begin{figure}[htb]
    \centering
    \includegraphics[width=1.0\textwidth]{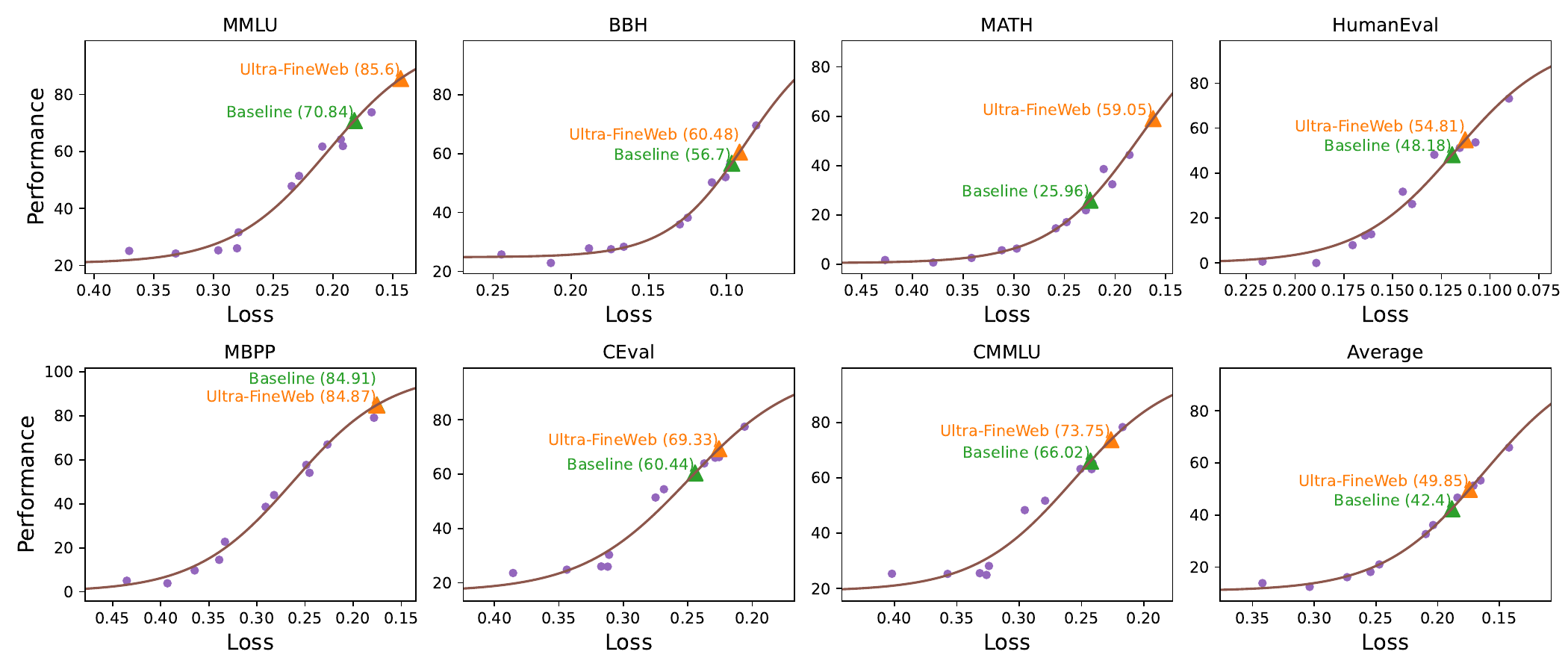}
    \caption{\textbf{Loss-performance curve:} Showing the estimated performance of an 8B model trained on 8T tokens using baseline and replacing high-quality data with Ultra-FineWeb.}
    \label{fig:densing_law}
\end{figure}

\textbf{Loss and Performance Estimation Results.}
We use the performance estimation methods proposed in~\citet{densing_law} for further analysis and verification of the effectiveness of Ultra-FineWeb. 
First, we establish the standard configuration in~\citet{densing_law} as the baseline. 
Specifically, we adopt the MiniCPM-3-4B~\citep{minicpm} training corpus, applying models across six scales (0.005B, 0.03B, 0.1B, 0.2B, 0.4B, 0.8B), and train with six token configurations ({10, 15, 20, 30, 40, 60} × N, where N represents the model parameter size).
Based on these 36 models, we compute and plot the compute ($=6ND$)-Loss curve, and subsequently predict the performance of each model using the Loss-Performance curve from the Densing Law. 
This analysis is performed on MMLU~\citep{mmlu}, BBH~\citep{bbh}, MATH~\citep{math}, MBPP~\citep{mbpp}, HumanEval~\citep{humaneval}, C-Eval~\citep{ceval}, and CMMLU~\citep{cmmlu} evaluation metrics. 
Next, we replace the “High-Quality” data in the baseline with \textbf{\textit{Ultra-FineWeb}} and repeat the experiment, performing Loss Estimation. 
Finally, through this two-step estimation, we predict the performance of an 8B model trained on 8T tokens. 
The loss values and estimated results are shown in Table~\ref{tab:densing_law}, with the Loss-Performance curve shown in Figure~\ref{fig:densing_law}. 
Experimental results demonstrate that using Ultra-FineWeb significantly reduces the loss for metrics such as MMLU, MATH, C-Eval, and CMMLU, thereby improves model performance.

\begin{table}[ht]
    \centering
    \small
    \captionsetup{justification=centerlast, singlelinecheck=false}
    \caption{Loss values and estimated performance for 8B model trained on 8T tokens.}
    \renewcommand{\arraystretch}{1.2}
    \begin{tabular}{lllll}
        \toprule
                                            
                          & \multicolumn{2}{c}{Baseline}     & \multicolumn{2}{c}{Ultra-FineWeb} \\
        \cmidrule(lr){2-3} \cmidrule(lr){4-5}
        Metrics            &  Loss	  &  Estimate Acc.	 &   Loss            	                    &  Estimate Acc.                                     \\
        \midrule
        MMLU	          &  0.182	  &  70.84	             &   0.143$_{\textcolor{red}{-0.039}}$	    &   \textbf{85.60}$_{\textcolor{red}{+14.76}}$   \\
        BBH	              &  0.097	  &  56.70	             &   0.092$_{\textcolor{red}{-0.005}}$	    &   \textbf{60.48}$_{\textcolor{red}{+3.78}}$    \\
        MATH	          &  0.225	  &  25.96	             &   0.162$_{\textcolor{red}{-0.063}}$	    &   \textbf{59.05}$_{\textcolor{red}{+33.09}}$   \\
        MBPP	          &  0.175	  &  \textbf{84.91}	     &   0.176$_{\textcolor{blue}{+0.001}}$	    &   84.87$_{\textcolor{blue}{-0.04}}$            \\
        HumanEval	      &  0.119	  &  48.18	             &   0.113$_{\textcolor{red}{-0.006}}$	    &   \textbf{54.81}$_{\textcolor{red}{+6.63}}$    \\
        C-Eval	          &  0.244	  &  60.44	             &   0.226$_{\textcolor{red}{-0.018}}$	    &   \textbf{69.33}$_{\textcolor{red}{+8.89}}$    \\
        CMMLU	          &  0.243	  &  66.02	             &   0.226$_{\textcolor{red}{-0.017}}$	    &   \textbf{73.75}$_{\textcolor{red}{+7.73}}$    \\
        \midrule
        \textit{Average}  &  0.189	  &  42.40	             &   0.174$_{\textcolor{red}{-0.015}}$	    &   \textbf{49.85}$_{\textcolor{red}{+7.45}}$    \\
        \bottomrule
    \end{tabular}
    \label{tab:densing_law}
\end{table}

\begin{figure}[htbp]
    \centering
    \begin{subfigure}{0.48\textwidth}
        \includegraphics[width=\linewidth]{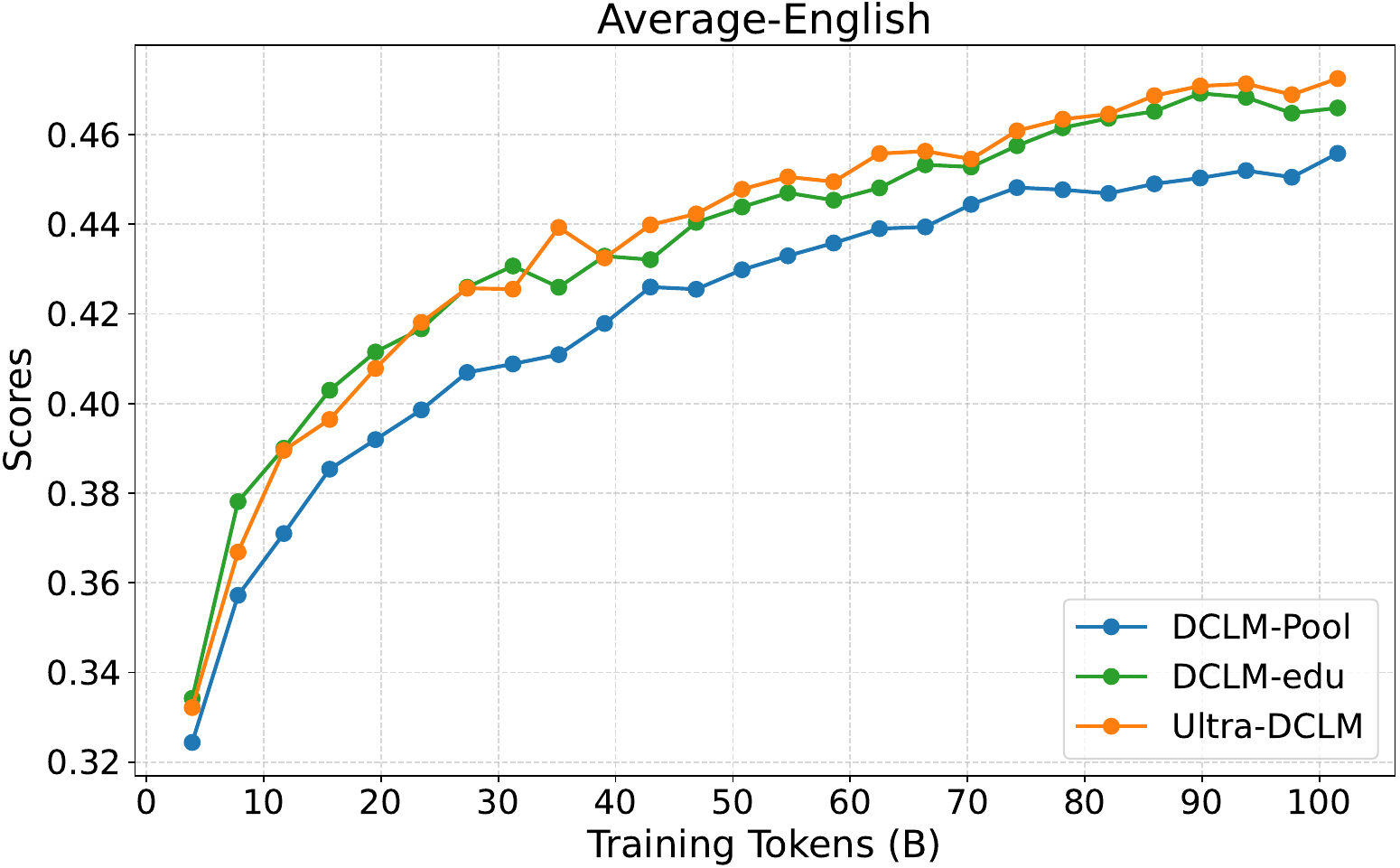}
        \captionsetup{justification=centerlast, singlelinecheck=false, width=0.95\linewidth}
        \caption{Comparison of results on DCLM-Pool-based datasets at each checkpoint.}
    \end{subfigure}
    \begin{subfigure}{0.48\textwidth}
        \includegraphics[width=\linewidth]{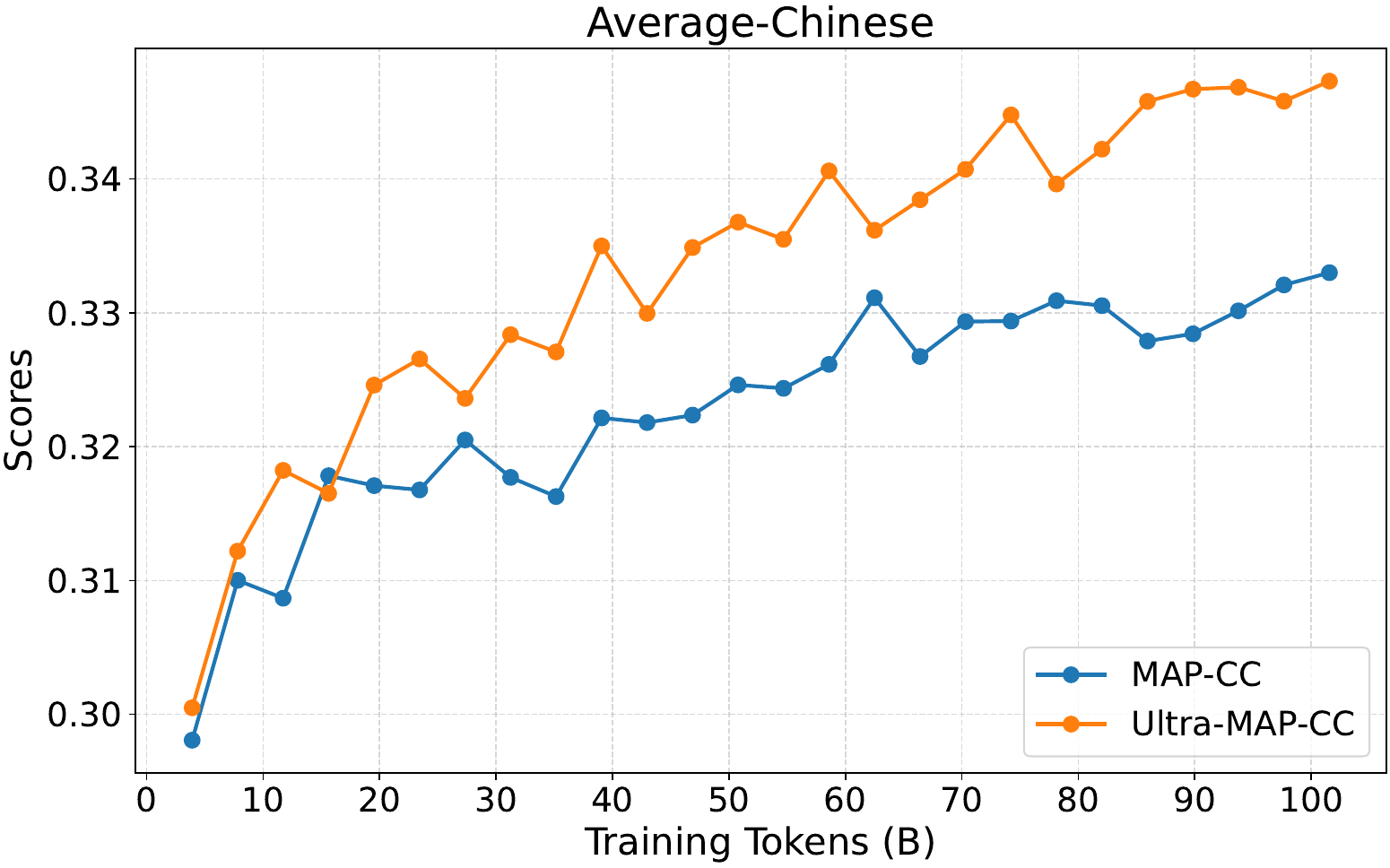}
        \captionsetup{justification=centerlast, singlelinecheck=false, width=0.95\linewidth}
        \caption{Comparison of results on MAP-CC-based datasets at each checkpoint.}
    \end{subfigure}
    \caption{Average scores at each checkpoint during training for different source data.}
    \label{fig:ab_multi_source}
\end{figure}

\textbf{Ablation Study on Multi-Source Seed Selection.}
To verify the impact of selecting multi-source seed on the robustness of the classifier during the efficient data filtering pipeline process, we choose DCLM-Pool~\citep{dclm} as the English data source and MAP-CC~\citep{mapcc} as the Chinese data source for verification. 
In the experiment, we compare the performance of models trained with original data, LLM-based classifier (-edu), and data extracted by our classifier (Ultra-) on different evaluation sets. 
Notably, due to the unavailability of an open-source LLM-based classifier for Chinese-FineWeb-edu, we only compare the performance difference between the original MAP-CC data and the data extracted by our classifier (Ultra-MAP-CC).
As detailed in Tables~\ref{tab:ab_dclm} and~\ref{tab:ab_mapcc}, Ultra-DCLM demonstrates superior performance over both DCLM-Pool and DCLM-edu across multiple English evaluation tasks. 
The English average score for Ultra-DCLM (47.252$pp$) shows a 1.671$pp$ improvement over DCLM-Pool (45.581$pp$) and a 0.658$pp$ advantage over DCLM-edu (46.594$pp$), with particularly notable gains in MMLU, ARC-C, and OpenbookQA metrics. 
For Chinese evaluations, Ultra-MAP-CC also exhibits significant enhancements, especially in CMMLU with a 2.8$pp$ increase, achieving an overall 1.43$pp$ improvement over the original dataset.
These results demonstrate that our classifier remains highly robust and effective even in non-homogeneous data scenarios, further confirming the positive impact of the multi-source seed selection strategy on improving classifier robustness and performance.
Figure~\ref{fig:ab_multi_source} presents the evaluation results at each checkpoint during training. In the early stages of training, the performance of Ultra-DCLM and DCLM-edu is similar, but both outperform DCLM-Pool significantly. 
When training reaches 30B tokens, Ultra-DCLM begins to surpass DCLM-edu. 
For the Chinese evaluation sets, Ultra-MAP-CC significantly outperforms MAP-CC from the early stages of training.

\begin{table}[h]
    \centering
    \small
    \caption{Comparison of results on DCLM-Pool-based datasets.}
    \renewcommand{\arraystretch}{1.2}
    \begin{tabular}{llll}
        \toprule
        Metrics                        & DCLM-Pool	     &   DCLM-edu		                               & Ultra-DCLM	                                      \\          
        \midrule
        MMLU	                       & 31.45	         &   34.07$_{\textcolor{red}{+2.62}}$	           & \textbf{34.33}$_{\textcolor{red}{+2.88}}$        \\
        ARC-C	                       & 31.48	         &   37.71$_{\textcolor{red}{+6.23}}$	           & \textbf{38.48}$_{\textcolor{red}{+7.00}}$        \\
        ARC-E	                       & 66.08	         &   \textbf{73.40}$_{\textcolor{red}{+7.32}}$	   & 72.77$_{\textcolor{red}{+6.69}}$                 \\
        CommonSenseQA	               & \textbf{41.52}	 &   39.72$_{\textcolor{blue}{-1.80}}$	           & 40.70$_{\textcolor{blue}{-0.82}}$                 \\
        HellaSwag	                   & \textbf{44.28}	 &   41.77$_{\textcolor{blue}{-2.51}}$	           & 43.31$_{\textcolor{blue}{-0.97}}$                 \\
        OpenbookQA	                   & 25.00	         &   26.60$_{\textcolor{red}{+1.60}}$	           & \textbf{27.40}$_{\textcolor{red}{+2.40}}$        \\
        PIQA	                       & 73.67	         &   70.73$_{\textcolor{blue}{-2.94}}$	           & \textbf{73.89}$_{\textcolor{red}{+0.22}}$        \\
        SIQA	                       & \textbf{40.79}	 &   39.00$_{\textcolor{blue}{-1.79}}$	           & 39.51$_{\textcolor{blue}{-1.28}}$                 \\
        Winogrande	                   & 55.96	         &   \textbf{56.35}$_{\textcolor{red}{+0.39}}$	   & 54.88$_{\textcolor{blue}{-1.08}}$                 \\
        \midrule
        \textit{Average$_{English}$}   & 45.581	         &   46.594$_{\textcolor{red}{+1.013}}$	           & \textbf{47.252}$_{\textcolor{red}{+1.671}}$      \\
        \bottomrule
    \end{tabular}
    \label{tab:ab_dclm}
\end{table} 

\begin{table}[h]
    \centering
    \small
    \caption{Comparison of results on MAP-CC-based datasets.}
    \renewcommand{\arraystretch}{1.2}
    \begin{tabular}{lll}
        \toprule
        Metrics                        & MAP-CC	        &   Ultra-MAP-CC	                                      \\
        \midrule
        C-Eval	                       & 34.58          & \textbf{34.64}$_{\textcolor{red}{+0.06}}$    \\
        CMMLU	                       & 32.02          & \textbf{34.82}$_{\textcolor{red}{+2.80}}$    \\
        \midrule
        \textit{Average$_{Chinese}$}   & 33.300	        &   \textbf{34.730}$_{\textcolor{red}{+1.430}}$ \\
        \bottomrule
    \end{tabular}
    \label{tab:ab_mapcc}
\end{table}

\textbf{Ablation Study of Multi-Turn Training Recipes.} 
To verify the impact of multiple iterations on classifier performance, we implement three rounds of iterations for both English and Chinese classifiers. 
The initial iteration utilizes the selected high-quality seed data for positive samples and multi-source original data for negative samples.
The second iteration involves using the classifier from the first round to process the negative samples, and the inferred positive and negative samples are incorporated into the next round of training data.
The third iteration involves updating the classifier with more precisely identified samples from the second round.
Experimental results (Tables~\ref{tab:ab_multi_turn_en} and ~\ref{tab:ab_multi_turn_zh}) indicate that second-iteration classifiers achieved superior performance across multiple tasks compared to the first-iteration. 
Notably, English classifiers demonstrate significant improvements in MMLU, ARC-C, and OpenbookQA tasks, with an average score increase of 3.613 percentage points ($pp$) over both the first iteration and original FineWeb dataset, reaching 45.89$pp$. 
However, the third iteration, which focused solely on updating samples from original source data, failed to yield additional performance gains. In fact, there were slight declines in some tasks, such as HellaSwag and PIQA.
For the Chinese data, the second iteration of Ultra-FineWeb-zh also shows notable improvements in CMMLU and C-Eval.
However, similar to the English results, the third iteration provided only marginal overall improvements, with no significant gains in specific tasks.
This suggests that iterative sample refinement through enhanced classifiers alone is insufficient for achieving further performance improvements.

\begin{table}[!htp]
    \centering
    \small
    \caption{Comparison of results on English datasets with multiple iterations.}
    \renewcommand{\arraystretch}{1.2}
    \begin{tabular}{lllll}
        \toprule
        Metrics	                          & FineWeb	      & fastText-en-v1		                             &  Ultra-FineWeb-en		                      & fastText-en-v3	                              \\
        \midrule
        MMLU	                          & 28.84	          & 32.30$_{\textcolor{red}{+3.46}}$	             &  32.24$_{\textcolor{red}{+3.40}}$	          & \textbf{32.29}$_{\textcolor{red}{+3.45}}$     \\
        ARC-C	                          & 25.17	          & \textbf{35.67}$_{\textcolor{red}{+10.50}}$	     &  \textbf{35.67}$_{\textcolor{red}{+10.50}}$	  & 35.07$_{\textcolor{red}{+9.9}}$               \\
        ARC-E	                          & 59.18	          & 70.33$_{\textcolor{red}{+11.15}}$	             &  \textbf{70.62}$_{\textcolor{red}{+11.44}}$	  & 70.54$_{\textcolor{red}{+11.36}}$             \\
        CommonSenseQA	                  & 34.32	          & 32.27$_{\textcolor{blue}{-2.05}}$	             &  36.45$_{\textcolor{red}{+2.13}}$	          & \textbf{36.55}$_{\textcolor{red}{+2.23}}$     \\
        HellaSwag	                      & \textbf{42.91}	  & 42.82$_{\textcolor{blue}{-0.09}}$	             &  42.76$_{\textcolor{blue}{-0.15}}$	          & 42.62$_{\textcolor{blue}{-0.29}}$             \\
        OpenbookQA	                      & 22.20	          & 24.40$_{\textcolor{red}{+2.20}}$	             &  \textbf{26.20}$_{\textcolor{red}{+4.00}}$	  & \textbf{26.20}$_{\textcolor{red}{+4.00}}$     \\
        PIQA	                          & 73.29	          & 72.09$_{\textcolor{blue}{-1.20}}$	             &  \textbf{73.67}$_{\textcolor{red}{+0.38}}$	  & 72.53$_{\textcolor{blue}{-0.76}}$             \\
        SIQA	                          & 38.95	          & 38.59$_{\textcolor{blue}{-0.36}}$	             &  \textbf{39.61}$_{\textcolor{red}{+0.66}}$	  & 39.41$_{\textcolor{red}{+0.46}}$              \\
        Winogrande	                      & 55.64	          & 55.09$_{\textcolor{blue}{-0.55}}$	             &  55.80$_{\textcolor{red}{+0.16}}$	          & \textbf{55.92}$_{\textcolor{red}{+0.28}}$     \\
        \midrule
        \textit{Average$_{English}$}	  & 42.278	          & 44.840$_{\textcolor{red}{+2.562}}$	             &  \textbf{45.891}$_{\textcolor{red}{+3.613}}$	  & 45.681$_{\textcolor{red}{+3.403}}$            \\
        \bottomrule
    \end{tabular}
    \label{tab:ab_multi_turn_en}
\end{table}

\begin{table}[!htp]
    \centering
    \small
    \caption{Comparison of results on Chinese datasets with multiple iterations.}
    \renewcommand{\arraystretch}{1.2}
    \begin{tabular}{lllll}
        \toprule
        Metrics	                        & Chinese-FineWeb	& fastText-zh-v1		                & Ultra-FineWeb-zh		                        & fastText-zh-v3	                        \\
        \midrule
        C-Eval	    	                & 33.95	            & 33.63$_{\textcolor{blue}{-0.32}}$ 	& \textbf{34.26}$_{\textcolor{red}{+0.31}}$ 	& \textbf{34.26}$_{\textcolor{red}{+0.31}}$ \\
        CMMLU	    	                & 32.41	            & 35.82$_{\textcolor{red}{+3.41}}$  	& \textbf{36.06}$_{\textcolor{red}{+3.65}}$	    & 35.07$_{\textcolor{red}{+2.66}}$          \\
        \midrule
        \textit{Average$_{Chinese}$}    & 34.035	        & 35.390$_{\textcolor{red}{+1.355}}$	& \textbf{35.875}$_{\textcolor{red}{+1.840}}$   & 34.26$_{\textcolor{red}{+0.225}}$         \\
        \bottomrule
    \end{tabular}
    \label{tab:ab_multi_turn_zh}
\end{table}

\begin{figure}[htbp]
    \centering
    \begin{subfigure}{0.48\textwidth}
        \includegraphics[width=\linewidth]{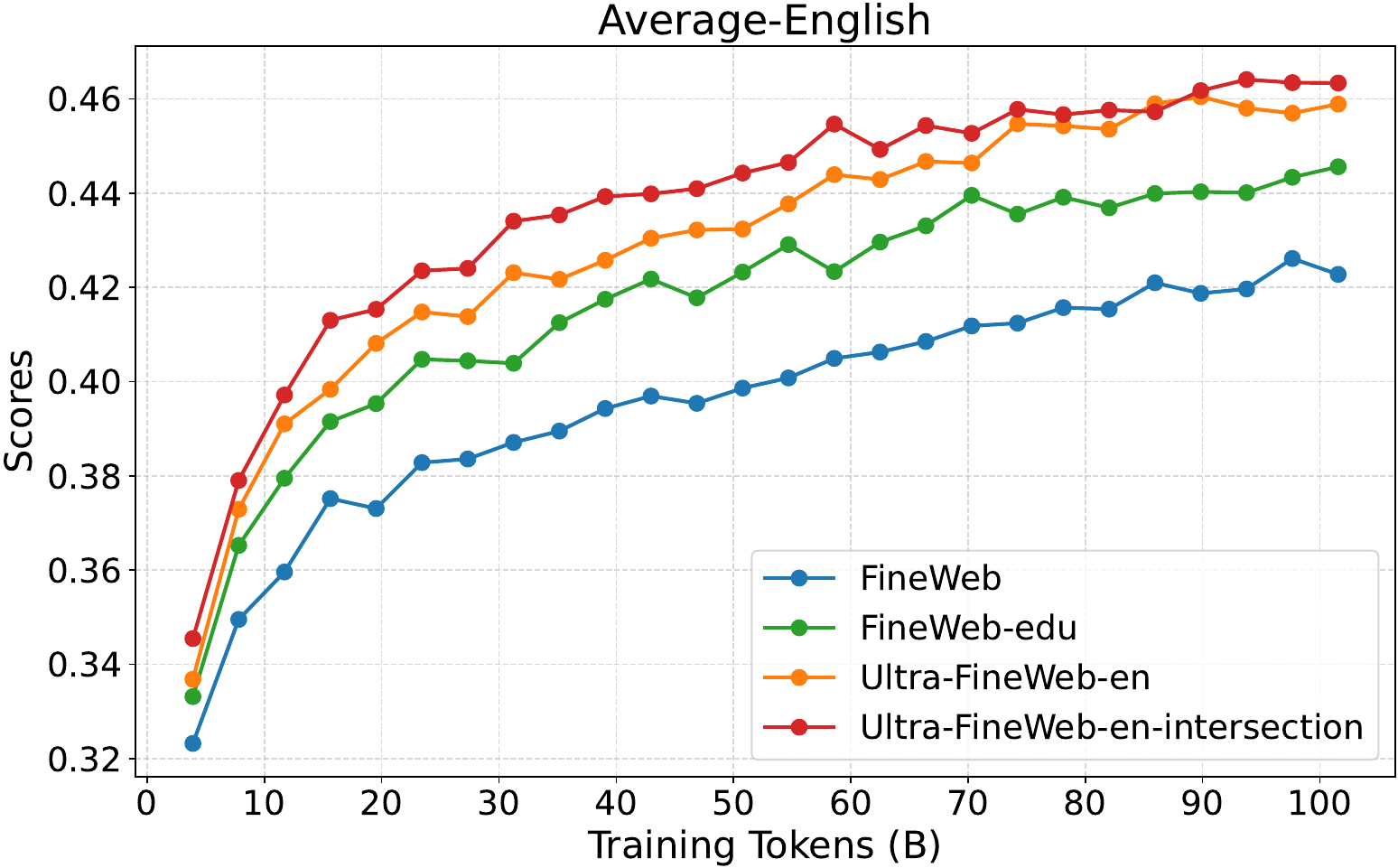}
        \captionsetup{justification=centerlast, singlelinecheck=false, width=0.9\linewidth}
        \caption{Comparison of results on English datasets.}
    \end{subfigure}
    \begin{subfigure}{0.48\textwidth}
        \includegraphics[width=\linewidth]{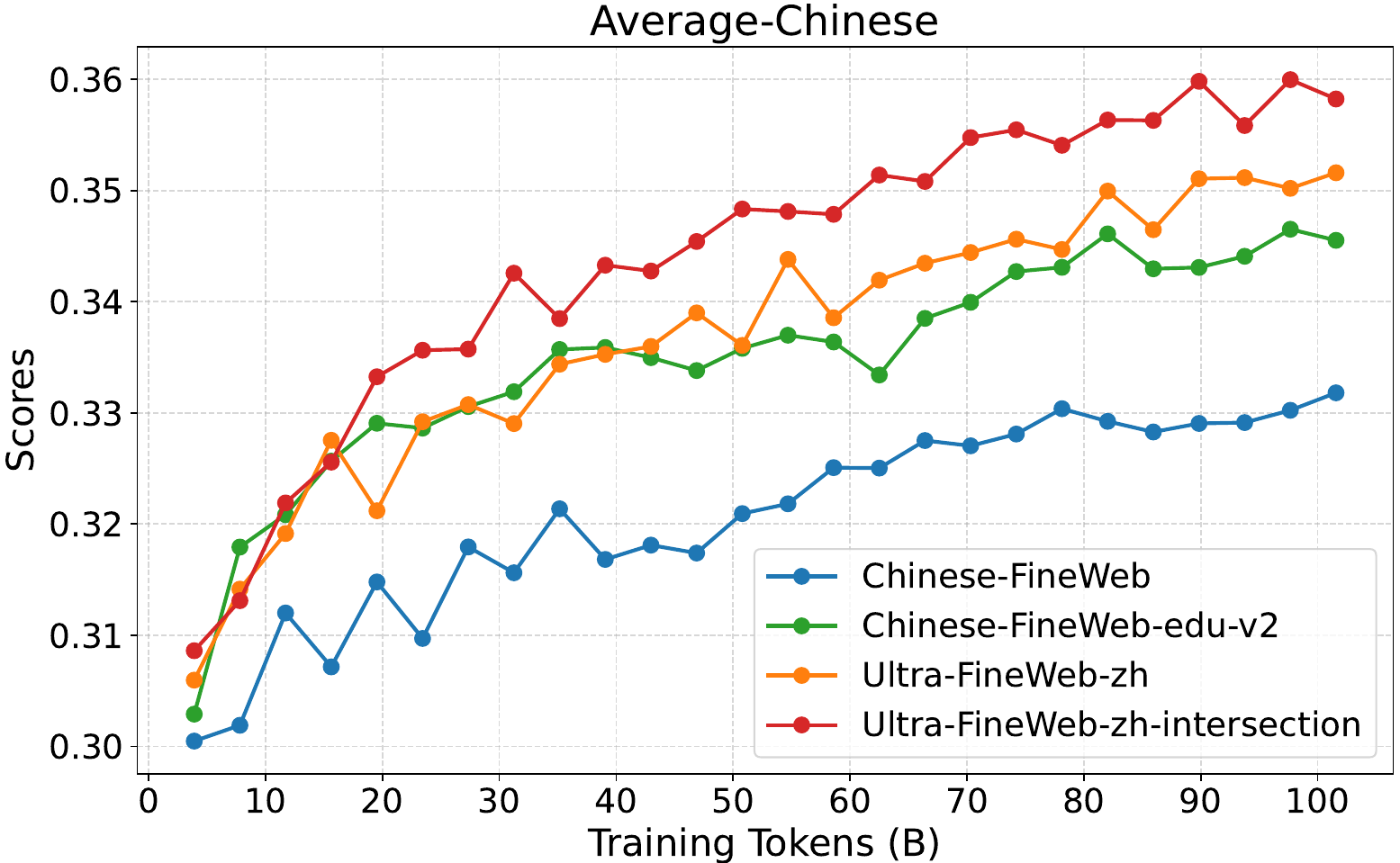}
        \captionsetup{justification=centerlast, singlelinecheck=false, width=0.9\linewidth}
        \caption{Comparison of results on Chinese datasets.}
    \end{subfigure}
    \caption{\textbf{Average scores at each checkpoint during training for different datasets}: Compared with using the intersection of positive samples inferred by multiple classifiers.}
    \label{fig:ab_intersection}
\end{figure}

\textbf{Ablation Study on Classifier Inference Intersection.}
To investigate the impact of intersecting positive samples from multiple classifiers on LLM performance, we conduct experiments using the intersection of classifier-inferred positive samples for model training.
As demonstrated in Tables~\ref{tab:ab_intersection_en} and~\ref{tab:ab_intersection_zh}, the model trained on Ultra-FineWeb-en$_{inter}$ exhibits substantial performance gains across multiple English metrics. 
Compared to the Ultra-FineWeb-en, the score improved by 0.447$pp$, with the most significant improvements observed in tasks such as MMLU, ARC-C, ARC-E, and OpenbookQA. 
Similarly, for Chinese metrics, the model trained on Ultra-FineWeb-zh$_{inter}$ also showed notable performance gains, with the overall Chinese average score increasing from 35.16$pp$ to 36.455$pp$.
In particular, the score in CMMLU improved by 1.8$pp$ compared to Ultra-FineWeb-zh. 
Additionally, we visualize the evaluation scores at each checkpoint during training, as shown in Figure~\ref{fig:ab_intersection}, where the model using intersection data consistently maintain the highest score throughout training. 
These results indicate that combining the inference results from multiple classifiers, particularly through intersecting positive sample data, can significantly further enhance model performance, yielding significant improvements across key metrics.

\begin{table}[htb]
    \centering
    \small
    \captionsetup{justification=centerlast, singlelinecheck=false}
    \caption{Comparison of results on English datasets using the intersection of positive samples inferred by multiple classifiers.}
    \renewcommand{\arraystretch}{1.2}
    \begin{tabular}{lllll}
        \toprule
        Metrics	                          & FineWeb	          &  FineWeb-edu		                         &   Ultra-FineWeb-en		                            &   Ultra-FineWeb-en$_{inter}$	                \\
        \midrule
        MMLU	                          & 28.84	          &  31.80$_{\textcolor{red}{+2.96}}$	         &   32.24$_{\textcolor{red}{+3.40}}$	                &   \textbf{33.37}$_{\textcolor{red}{+4.53}}$       \\
        ARC-C	                          & 25.17	          &  34.56$_{\textcolor{red}{+9.39}}$	         &   35.67$_{\textcolor{red}{+10.50}}$	                &   \textbf{38.31}$_{\textcolor{red}{+13.14}}$      \\
        ARC-E	                          & 59.18	          &  69.95$_{\textcolor{red}{+10.77}}$	         &   70.62$_{\textcolor{red}{+11.44}}$	                &   \textbf{73.48}$_{\textcolor{red}{+14.30}}$      \\
        CommonSenseQA	                  & 34.32	          &  31.53$_{\textcolor{blue}{-2.79}}$	         &   36.45$_{\textcolor{red}{+2.13}}$	                &   \textbf{36.94}$_{\textcolor{red}{+2.62}}$       \\
        HellaSwag	                      & \textbf{42.91}	  &  42.17$_{\textcolor{blue}{-0.74}}$	         &   42.76$_{\textcolor{blue}{-0.15}}$	                &   41.39$_{\textcolor{blue}{-1.52}}$               \\
        OpenbookQA	                      & 22.20	          &  25.20$_{\textcolor{red}{+3.00}}$	         &   26.20$_{\textcolor{red}{+4.00}}$	                &   \textbf{28.60}$_{\textcolor{red}{+6.40}}$       \\
        PIQA	                          & 73.29	          &  72.14$_{\textcolor{blue}{-1.15}}$	         &   \textbf{73.67}$_{\textcolor{red}{+0.38}}$	        &   71.16$_{\textcolor{blue}{-2.13}}$               \\
        SIQA	                          & 38.95	          &  38.13$_{\textcolor{blue}{-0.82}}$	         &   \textbf{39.61}$_{\textcolor{red}{+0.66}}$	        &   39.41$_{\textcolor{red}{+0.46}}$                \\
        Winogrande	                      & 55.64	          &  55.56$_{\textcolor{blue}{-0.08}}$	         &   \textbf{55.80}$_{\textcolor{red}{+0.16}}$	        &   54.38$_{\textcolor{blue}{-1.26}}$               \\
        \midrule
        \textit{Average$_{English}$}	  & 42.278	          &  44.560$_{\textcolor{red}{+2.282}}$	         &   45.891$_{\textcolor{red}{+3.613}}$	                &   \textbf{46.338}$_{\textcolor{red}{+4.06}}$      \\
        \bottomrule
    \end{tabular}
    \label{tab:ab_intersection_en}
\end{table}

\begin{table}[htb]
    \centering
    \small
    \captionsetup{justification=centerlast, singlelinecheck=false}
    \caption{Comparison of results on Chinese Datasets using the intersection of positive samples inferred by multiple classifiers.}
    \renewcommand{\arraystretch}{1.2}
    \begin{tabular}{lllll}
        \toprule
        Metrics	                       & Chinese-FineWeb	    & Chinese-FineWeb-edu-v2		                  &  Ultra-FineWeb-zh		                              &  Ultra-FineWeb-zh$_{inter}$                      \\	
        \midrule
        C-Eval	                       & 33.95	                & 34.17$_{\textcolor{red}{+0.22}}$	              &  34.26$_{\textcolor{red}{+0.31}}$	                  & \textbf{35.05}$_{\textcolor{red}{+1.1}}$        \\
        CMMLU	                       & 32.41	                & 34.93$_{\textcolor{red}{+2.52}}$	              &  36.06$_{\textcolor{red}{+3.65}}$	                  & \textbf{37.86}$_{\textcolor{red}{+5.45}}$       \\
        \midrule
        \textit{Average$_{Chinese}$}   & 33.180	                & 34.550$_{\textcolor{red}{+1.37}}$	              &  35.160$_{\textcolor{red}{+1.98}}$	                  & \textbf{36.455}$_{\textcolor{red}{+3.275}}$     \\
        \bottomrule
    \end{tabular}
    \label{tab:ab_intersection_zh}
\end{table}

\vspace{-1.0em}
\section{Related Work}
\vspace{-1.0em}
The success of LLMs largely depends on the availability of large-scale, high-quality pretraining corpora, which provide models with rich knowledge and reasoning capabilities. 
Common Crawl~\citep{common_crawl} has served as the foundation data source for LLM development, and to meet the growing data requirements for training larger models, a vast amount of pretraining corpora have been made open source.
Early efforts, such as C4~\citep{c4} with 160B tokens and Pile~\citep{pile} with 300B tokens, provide critical resources for early model pretraining. 
In recent years, substantially larger corpora have emerged, including RefinedWeb~\citep{refineweb} with 600B tokens, Dolma~\citep{dolma} with 3T tokens, FineWeb~\citep{fineweb} with 15T tokens, RedPajama-v2~\citep{redpajama} with 30T tokens, and DCLM~\citep{dclm} with 240T tokens, significantly advancing LLM development, fostering community collaboration, and establishing new benchmarks for innovation. 
Meanwhile, Chinese pretraining corpora have also been rapidly developed, such as ChineseWebText~\citep{chinesewebtext} with 50B tokens, WuDao~\citep{wudao} with 120B tokens, IndustryCorpus2~\citep{industryCorpus2} with 200B tokens, and CCI3~\citep{cci3} with 200B tokens. 
However, despite progress in traditional data processing methods (such as heuristic filtering and deduplication) during the early stages, the processed data still often contains noise and unstructured content.
With the continuous scaling up of models and increasing demands for data quality, these methods have become insufficient to meet current requirements.

To address these challenges, model-driven data filtering strategies have gradually become an effective approach to improving data quality in recent years. 
These approaches are primarily implemented during the final stages of large-scale data preprocessing, aiming to filter high-quality and high-value samples from massive datasets to further enhance model performance. 
Traditional quality filtering techniques~\citep{fineweb, cci3, dclm, chinese_fineweb} typically train classifiers to distinguish between high-quality data (such as textbook text) and low-quality data (such as raw web text), subsequently filtering out samples with lower inference scores. 
Additionally, data filtering methods based on perplexity~\citep{scaling_data, ccnet}, and strategies using pre-trained LLMs to evaluate multiple dimensions of data quality through prompts~\citep{how_to, qurating}, have been introduced.
These advancements have greatly expanded the range of data filtering methods available.

The common trend of these methods is to obtain higher-quality data by reducing computational costs. 
By optimizing the filtering process and reducing inference resource consumption, not only is dataset quality improved, but data processing efficiency is also accelerated. 
This optimization enables LLMs to access superior training corpora, facilitating enhanced model performance with reduced training token requirements.

\vspace{-1.0em}
\section{Conclusion}
\vspace{-1.0em}
In this paper, we construct a higher-quality \textbf{\textit{Ultra-FineWeb}} dataset (including English data \textit{Ultra-FineWeb-en}, approximately 1T tokens, and Chinese data \textit{Ultra-FineWeb-zh}, approximately 120B tokens, totaling approximately 1.1T tokens).
This dataset is based on the FineWeb and Chinese FineWeb datasets, utilizing our proposed efficient data filtering pipeline.
Through rigorous experimental evaluations, we demonstrate that Ultra-FineWeb-en and Ultra-FineWeb-zh outperform FineWeb-edu and Chinese FineWeb-edu-v2 when used for small-scale model training from scratch.
Additionally, we show the effectiveness of the high-quality data filtered by our classifier on the DCLM-Pool and MAP-CC datasets, further confirming the reliability and effectiveness of our proposed pipeline. 
These results indicate that classifiers based on our efficient data filtering pipeline can select higher-quality data with reduced computational cost, thereby improving model training performance.
We provide a detailed description of the implementation of our efficient data filtering pipeline, especially the efficient verification strategy driven by classifiers in the pipeline. 
This strategy enables reliable assessment of training data impact on LLM performance while maintaining minimal computational requirements.
Furthermore, we present detailed methodologies for classifier seed data selection, training recipes, and FastText model training configuration, ensuring experimental reproducibility and result transparency.
This study aims to provide novel insights and methodologies for high-quality data filtering, offering valuable references for data quality optimization in future LLM training processes, and contributing to the further development of LLMs.

\vspace{-1.0em}
\section{Limitations and Future Directions}
\vspace{-1.0em}
Some key limitations of our work are as follows. Due to time and resource constraints, we did not conduct an comprehensive analysis of classifier inference thresholds.
Although the default threshold ($thr=0.5$) is effective in filtering higher-quality data, in future work, we plan to explore the impact of different threshold ranges on data quality to further optimize the filtering strategy. 
Specifically, during multiple iterations, we can experiment with different thresholds to refine the filtering process, enabling more precise selection of data at different quality levels for the next round of classifier training. 
This approach will help dynamically adjust the filtering strategy and progressively optimize the dataset, thereby improving classifier performance and data quality.
Additionally, this paper primarily focuses on general high-quality datasets and validates the effectiveness of the efficient data filtering pipeline. 
In the future, we hope to extend this method to more specialized domains, such as mathematics, code, law, and other technical fields, to meet the requirements of different scenarios. 
This would contribute to the creation of more targeted high-quality datasets, enhancing model performance on specific tasks.
Furthermore, current work typically evaluates data quality through model training results, which is heavily reliant on the performance of the trained models and lacks more objective and systematic quality metrics. 
Therefore, in future research, we aim to develop some quantifiable data quality evaluation standards or tools to provide multidimensional measurements of data quality. 
This would further enhance the precision and operability of data filtering, providing a more scientific basis for building high-quality datasets. 
These efforts will further improve our data filtering methods and provide higher-quality data support for the training of large-scale language models.

\newpage

\bibliographystyle{citation}
\bibliography{citation}

\begin{thebibliography}{63}
\providecommand{\natexlab}[1]{#1}
\providecommand{\url}[1]{\texttt{#1}}
\expandafter\ifx\csname urlstyle\endcsname\relax
  \providecommand{\doi}[1]{doi: #1}\else
  \providecommand{\doi}{doi: \begingroup \urlstyle{rm}\Url}\fi

\bibitem[Austin et~al.(2021)Austin, Odena, Nye, Bosma, Michalewski, Dohan, Jiang, Cai, Terry, Le, et~al.]{mbpp}
Jacob Austin, Augustus Odena, Maxwell Nye, Maarten Bosma, Henryk Michalewski, David Dohan, Ellen Jiang, Carrie Cai, Michael Terry, Quoc Le, et~al.
\newblock Program synthesis with large language models.
\newblock \emph{arXiv preprint arXiv:2108.07732}, 2021.

\bibitem[{BAAI}(2023)]{wudao}
{BAAI}.
\newblock Wudao corpus, 2023.
\newblock URL \url{https://data.baai.ac.cn/details/WuDaoCorporaText}.

\bibitem[Bai et~al.(2023)Bai, Bai, Chu, Cui, Dang, Deng, Fan, Ge, Han, Huang, et~al.]{qwen_tech_report}
Jinze Bai, Shuai Bai, Yunfei Chu, Zeyu Cui, Kai Dang, Xiaodong Deng, Yang Fan, Wenbin Ge, Yu~Han, Fei Huang, et~al.
\newblock Qwen technical report.
\newblock \emph{arXiv preprint arXiv:2309.16609}, 2023.

\bibitem[Bisk et~al.(2020)Bisk, Zellers, Gao, Choi, et~al.]{piqa}
Yonatan Bisk, Rowan Zellers, Jianfeng Gao, Yejin Choi, et~al.
\newblock Piqa: Reasoning about physical commonsense in natural language.
\newblock In \emph{Proceedings of the AAAI conference on artificial intelligence}, volume~34, pp.\  7432--7439, 2020.

\bibitem[Cai et~al.(2024)Cai, Cao, Chen, Chen, Chen, Chen, Chen, Chen, Chen, Chu, et~al.]{internlm2}
Zheng Cai, Maosong Cao, Haojiong Chen, Kai Chen, Keyu Chen, Xin Chen, Xun Chen, Zehui Chen, Zhi Chen, Pei Chu, et~al.
\newblock Internlm2 technical report.
\newblock \emph{arXiv preprint arXiv:2403.17297}, 2024.

\bibitem[Chen et~al.(2023)Chen, Jian, Xi, Yi, Du, Ding, Zhu, Zong, Wang, and Zhang]{chinesewebtext}
Jianghao Chen, Pu~Jian, Tengxiao Xi, Dongyi Yi, Qianlong Du, Chenglin Ding, Guibo Zhu, Chengqing Zong, Jinqiao Wang, and Jiajun Zhang.
\newblock {ChineseWebText}: Large-scale high-quality chinese web text extracted with effective evaluation model.
\newblock \emph{arXiv preprint arXiv:2311.01149}, 2023.

\bibitem[Chen et~al.(2021)Chen, Tworek, Jun, Yuan, Pinto, Kaplan, Edwards, Burda, Joseph, Brockman, et~al.]{humaneval}
Mark Chen, Jerry Tworek, Heewoo Jun, Qiming Yuan, Henrique Ponde De~Oliveira Pinto, Jared Kaplan, Harri Edwards, Yuri Burda, Nicholas Joseph, Greg Brockman, et~al.
\newblock Evaluating large language models trained on code.
\newblock \emph{arXiv preprint arXiv:2107.03374}, 2021.

\bibitem[Clark et~al.(2018)Clark, Cowhey, Etzioni, Khot, Sabharwal, Schoenick, and Tafjord]{arc}
Peter Clark, Isaac Cowhey, Oren Etzioni, Tushar Khot, Ashish Sabharwal, Carissa Schoenick, and Oyvind Tafjord.
\newblock Think you have solved question answering? try arc, the ai2 reasoning challenge.
\newblock \emph{arXiv preprint arXiv:1803.05457}, 2018.

\bibitem[Contributors(2023)]{opencompass}
OpenCompass Contributors.
\newblock Opencompass: A universal evaluation platform for foundation models.
\newblock \url{https://github.com/open-compass/opencompass}, 2023.

\bibitem[Crawl(2007)]{common_crawl}
Common Crawl.
\newblock Common crawl.
\newblock \url{https://commoncrawl.org}, 2007.

\bibitem[Du et~al.(2024)Du, Yu, Gao, Pan, Cheng, Ma, Yuan, Qu, Liu, Zheng, et~al.]{mapcc}
Xinrun Du, Zhouliang Yu, Songyang Gao, Ding Pan, Yuyang Cheng, Ziyang Ma, Ruibin Yuan, Xingwei Qu, Jiaheng Liu, Tianyu Zheng, et~al.
\newblock Chinese tiny llm: Pretraining a chinese-centric large language model.
\newblock \emph{arXiv preprint arXiv:2404.04167}, 2024.

\bibitem[Dubey et~al.(2024)Dubey, Jauhri, Pandey, Kadian, Al-Dahle, Letman, Mathur, Schelten, Yang, Fan, et~al.]{llama3}
Abhimanyu Dubey, Abhinav Jauhri, Abhinav Pandey, Abhishek Kadian, Ahmad Al-Dahle, Aiesha Letman, Akhil Mathur, Alan Schelten, Amy Yang, Angela Fan, et~al.
\newblock The llama 3 herd of models.
\newblock \emph{arXiv preprint arXiv:2407.21783}, 2024.

\bibitem[Fourrier et~al.(2023)Fourrier, Habib, Kydlíček, Wolf, and Tunstall]{lighteval}
Clémentine Fourrier, Nathan Habib, Hynek Kydlíček, Thomas Wolf, and Lewis Tunstall.
\newblock Lighteval: A lightweight framework for llm evaluation, 2023.
\newblock URL \url{https://github.com/huggingface/lighteval}.

\bibitem[Gao et~al.(2020)Gao, Biderman, Black, Golding, Hoppe, Foster, Phang, He, Thite, Nabeshima, et~al.]{pile}
Leo Gao, Stella Biderman, Sid Black, Laurence Golding, Travis Hoppe, Charles Foster, Jason Phang, Horace He, Anish Thite, Noa Nabeshima, et~al.
\newblock The {P}ile: An 800{GB} dataset of diverse text for language modeling.
\newblock \emph{arXiv preprint arXiv:2101.00027}, 2020.

\bibitem[Gunasekar et~al.(2023)Gunasekar, Zhang, Aneja, Mendes, Del~Giorno, Gopi, Javaheripi, Kauffmann, de~Rosa, Saarikivi, et~al.]{phi1}
Suriya Gunasekar, Yi~Zhang, Jyoti Aneja, Caio C{\'e}sar~Teodoro Mendes, Allie Del~Giorno, Sivakanth Gopi, Mojan Javaheripi, Piero Kauffmann, Gustavo de~Rosa, Olli Saarikivi, et~al.
\newblock Textbooks are all you need.
\newblock \emph{arXiv preprint arXiv:2306.11644}, 2023.

\bibitem[Guo et~al.(2024)Guo, Zhu, Yang, Xie, Dong, Zhang, Chen, Bi, Wu, Li, et~al.]{deepseekcoder}
Daya Guo, Qihao Zhu, Dejian Yang, Zhenda Xie, Kai Dong, Wentao Zhang, Guanting Chen, Xiao Bi, Yu~Wu, YK~Li, et~al.
\newblock Deepseek-coder: When the large language model meets programming--the rise of code intelligence.
\newblock \emph{arXiv preprint arXiv:2401.14196}, 2024.

\bibitem[Guo et~al.(2025)Guo, Yang, Zhang, Song, Zhang, Xu, Zhu, Ma, Wang, Bi, et~al.]{deepseekr1}
Daya Guo, Dejian Yang, Haowei Zhang, Junxiao Song, Ruoyu Zhang, Runxin Xu, Qihao Zhu, Shirong Ma, Peiyi Wang, Xiao Bi, et~al.
\newblock Deepseek-r1: Incentivizing reasoning capability in llms via reinforcement learning.
\newblock \emph{arXiv preprint arXiv:2501.12948}, 2025.

\bibitem[Han et~al.(2021)Han, Zhang, Ding, Gu, Liu, Huo, Qiu, Yao, Zhang, Zhang, et~al.]{han2021pre}
Xu~Han, Zhengyan Zhang, Ning Ding, Yuxian Gu, Xiao Liu, Yuqi Huo, Jiezhong Qiu, Yuan Yao, Ao~Zhang, Liang Zhang, et~al.
\newblock Pre-trained models: Past, present and future.
\newblock \emph{AI Open}, 2:\penalty0 225--250, 2021.

\bibitem[He et~al.(2024)He, Wang, Liu, Liu, Yao, Huang, Li, Li, Che, Zhang, et~al.]{telechat}
Zhongjiang He, Zihan Wang, Xinzhang Liu, Shixuan Liu, Yitong Yao, Yuyao Huang, Xuelong Li, Yongxiang Li, Zhonghao Che, Zhaoxi Zhang, et~al.
\newblock Telechat technical report.
\newblock \emph{arXiv preprint arXiv:2401.03804}, 2024.

\bibitem[Hendrycks et~al.(2020)Hendrycks, Burns, Basart, Zou, Mazeika, Song, and Steinhardt]{mmlu}
Dan Hendrycks, Collin Burns, Steven Basart, Andy Zou, Mantas Mazeika, Dawn Song, and Jacob Steinhardt.
\newblock Measuring massive multitask language understanding.
\newblock \emph{arXiv preprint arXiv:2009.03300}, 2020.

\bibitem[Hendrycks et~al.(2021)Hendrycks, Burns, Kadavath, Arora, Basart, Tang, Song, and Steinhardt]{math}
Dan Hendrycks, Collin Burns, Saurav Kadavath, Akul Arora, Steven Basart, Eric Tang, Dawn Song, and Jacob Steinhardt.
\newblock Measuring mathematical problem solving with the math dataset.
\newblock \emph{arXiv preprint arXiv:2103.03874}, 2021.

\bibitem[Hu et~al.(2024)Hu, Tu, Han, He, Cui, Long, Zheng, Fang, Huang, Zhao, et~al.]{minicpm}
Shengding Hu, Yuge Tu, Xu~Han, Chaoqun He, Ganqu Cui, Xiang Long, Zhi Zheng, Yewei Fang, Yuxiang Huang, Weilin Zhao, et~al.
\newblock {MiniCPM}: Unveiling the potential of small language models with scalable training strategies.
\newblock \emph{arXiv preprint arXiv:2404.06395}, 2024.

\bibitem[Huang et~al.(2023)Huang, Bai, Zhu, Zhang, Zhang, Su, Liu, Lv, Zhang, Lei, Fu, Sun, and He]{ceval}
Yuzhen Huang, Yuzhuo Bai, Zhihao Zhu, Junlei Zhang, Jinghan Zhang, Tangjun Su, Junteng Liu, Chuancheng Lv, Yikai Zhang, Jiayi Lei, Yao Fu, Maosong Sun, and Junxian He.
\newblock C-eval: A multi-level multi-discipline chinese evaluation suite for foundation models.
\newblock \emph{arXiv preprint arXiv:2305.08322}, 2023.

\bibitem[Joulin et~al.(2016)Joulin, Grave, Bojanowski, Douze, J{\'e}gou, and Mikolov]{fasttext}
Armand Joulin, Edouard Grave, Piotr Bojanowski, Matthijs Douze, H{\'e}rve J{\'e}gou, and Tomas Mikolov.
\newblock {FastText.zip}: Compressing text classification models.
\newblock \emph{arXiv preprint arXiv:1612.03651}, 2016.

\bibitem[Lee et~al.(2021)Lee, Ippolito, Nystrom, Zhang, Eck, Callison-Burch, and Carlini]{deduplicating}
Katherine Lee, Daphne Ippolito, Andrew Nystrom, Chiyuan Zhang, Douglas Eck, Chris Callison-Burch, and Nicholas Carlini.
\newblock Deduplicating training data makes language models better.
\newblock \emph{arXiv preprint arXiv:2107.06499}, 2021.

\bibitem[Li et~al.(2023)Li, Zhang, Koto, Yang, Zhao, Gong, Duan, and Baldwin]{cmmlu}
Haonan Li, Yixuan Zhang, Fajri Koto, Yifei Yang, Hai Zhao, Yeyun Gong, Nan Duan, and Timothy Baldwin.
\newblock Cmmlu: Measuring massive multitask language understanding in chinese, 2023.

\bibitem[Li et~al.(2024)Li, Fang, Smyrnis, Ivgi, Jordan, Gadre, Bansal, Guha, Keh, Arora, et~al.]{dclm}
Jeffrey Li, Alex Fang, Georgios Smyrnis, Maor Ivgi, Matt Jordan, Samir Gadre, Hritik Bansal, Etash Guha, Sedrick Keh, Kushal Arora, et~al.
\newblock Datacomp-lm: In search of the next generation of training sets for language models.
\newblock \emph{arXiv preprint arXiv:2406.11794}, 2024.

\bibitem[Liu et~al.(2024)Liu, Feng, Wang, Wang, Liu, Zhao, Dengr, Ruan, Dai, Guo, et~al.]{deepseekv2}
Aixin Liu, Bei Feng, Bin Wang, Bingxuan Wang, Bo~Liu, Chenggang Zhao, Chengqi Dengr, Chong Ruan, Damai Dai, Daya Guo, et~al.
\newblock Deepseek-v2: A strong, economical, and efficient mixture-of-experts language model.
\newblock \emph{arXiv preprint arXiv:2405.04434}, 2024.

\bibitem[Liu et~al.(2023)Liu, Shang, Wang, Xu, Wang, Li, Li, and He]{michao}
Yidong Liu, FuKai Shang, Fang Wang, Rui Xu, Jun Wang, Wei Li, Yao Li, and Conghui He.
\newblock Michao-huafen 1.0: A specialized pre-trained corpus dataset for domain-specific large models.
\newblock \emph{arXiv preprint arXiv:2309.13079}, 2023.

\bibitem[Lozhkov et~al.(2024)Lozhkov, Li, Allal, Cassano, Lamy-Poirier, Tazi, Tang, Pykhtar, Liu, Wei, et~al.]{starcoder2}
Anton Lozhkov, Raymond Li, Loubna~Ben Allal, Federico Cassano, Joel Lamy-Poirier, Nouamane Tazi, Ao~Tang, Dmytro Pykhtar, Jiawei Liu, Yuxiang Wei, et~al.
\newblock Starcoder 2 and the stack v2: The next generation.
\newblock \emph{arXiv preprint arXiv:2402.19173}, 2024.

\bibitem[Luo et~al.(2025)Luo, Yang, Xu, Yang, and Du]{LLM4SR}
Ziming Luo, Zonglin Yang, Zexin Xu, Wei Yang, and Xinya Du.
\newblock Llm4sr: A survey on large language models for scientific research.
\newblock \emph{arXiv preprint arXiv:2501.04306}, 2025.

\bibitem[Lyu et~al.(2025)Lyu, Gao, Gu, Zhang, Gao, Liu, Wang, Li, Zhao, Huang, et~al.]{OREAL}
Chengqi Lyu, Songyang Gao, Yuzhe Gu, Wenwei Zhang, Jianfei Gao, Kuikun Liu, Ziyi Wang, Shuaibin Li, Qian Zhao, Haian Huang, et~al.
\newblock Exploring the limit of outcome reward for learning mathematical reasoning.
\newblock \emph{arXiv preprint arXiv:2502.06781}, 2025.

\bibitem[Mihaylov et~al.(2018)Mihaylov, Clark, Khot, and Sabharwal]{openbookqa}
Todor Mihaylov, Peter Clark, Tushar Khot, and Ashish Sabharwal.
\newblock Can a suit of armor conduct electricity? a new dataset for open book question answering.
\newblock \emph{arXiv preprint arXiv:1809.02789}, 2018.

\bibitem[Muennighoff et~al.(2024)Muennighoff, Rush, Barak, Le~Scao, Tazi, Piktus, Pyysalo, Wolf, and Raffel]{scaling_data}
Niklas Muennighoff, Alexander Rush, Boaz Barak, Teven Le~Scao, Nouamane Tazi, Aleksandra Piktus, Sampo Pyysalo, Thomas Wolf, and Colin~A Raffel.
\newblock Scaling data-constrained language models.
\newblock \emph{Advances in Neural Information Processing Systems}, 36, 2024.

\bibitem[Ouyang et~al.(2022)Ouyang, Wu, Jiang, Almeida, Wainwright, Mishkin, Zhang, Agarwal, Slama, Ray, et~al.]{instructGPT}
Long Ouyang, Jeffrey Wu, Xu~Jiang, Diogo Almeida, Carroll Wainwright, Pamela Mishkin, Chong Zhang, Sandhini Agarwal, Katarina Slama, Alex Ray, et~al.
\newblock Training language models to follow instructions with human feedback.
\newblock \emph{Advances in neural information processing systems}, 35:\penalty0 27730--27744, 2022.

\bibitem[Penedo et~al.(2023)Penedo, Malartic, Hesslow, Cojocaru, Cappelli, Alobeidli, Pannier, Almazrouei, and Launay]{refineweb}
Guilherme Penedo, Quentin Malartic, Daniel Hesslow, Ruxandra Cojocaru, Alessandro Cappelli, Hamza Alobeidli, Baptiste Pannier, Ebtesam Almazrouei, and Julien Launay.
\newblock The refinedweb dataset for falcon llm: outperforming curated corpora with web data, and web data only.
\newblock \emph{arXiv preprint arXiv:2306.01116}, 2023.

\bibitem[Penedo et~al.(2024)Penedo, Kydl{\'\i}{\v{c}}ek, Lozhkov, Mitchell, Raffel, Von~Werra, Wolf, et~al.]{fineweb}
Guilherme Penedo, Hynek Kydl{\'\i}{\v{c}}ek, Anton Lozhkov, Margaret Mitchell, Colin Raffel, Leandro Von~Werra, Thomas Wolf, et~al.
\newblock The fineweb datasets: Decanting the web for the finest text data at scale.
\newblock \emph{arXiv preprint arXiv:2406.17557}, 2024.

\bibitem[Qiu et~al.(2024)Qiu, Lv, Jin, Wang, Ning, Yu, Zhang, Li, Chu, Qu, et~al.]{wanjuan}
Jiantao Qiu, Haijun Lv, Zhenjiang Jin, Rui Wang, Wenchang Ning, Jia Yu, ChaoBin Zhang, Zhenxiang Li, Pei Chu, Yuan Qu, et~al.
\newblock {Wanjuan-cc}: A safe and high-quality open-sourced english webtext dataset.
\newblock \emph{arXiv preprint arXiv:2402.19282}, 2024.

\bibitem[Rae et~al.(2021)Rae, Borgeaud, Cai, Millican, Hoffmann, Song, Aslanides, Henderson, Ring, Young, et~al.]{rae2021scaling}
Jack~W Rae, Sebastian Borgeaud, Trevor Cai, Katie Millican, Jordan Hoffmann, Francis Song, John Aslanides, Sarah Henderson, Roman Ring, Susannah Young, et~al.
\newblock Scaling language models: Methods, analysis \& insights from training gopher.
\newblock \emph{arXiv preprint arXiv:2112.11446}, 2021.

\bibitem[Raffel et~al.(2020)Raffel, Shazeer, Roberts, Lee, Narang, Matena, Zhou, Li, and Liu]{c4}
Colin Raffel, Noam Shazeer, Adam Roberts, Katherine Lee, Sharan Narang, Michael Matena, Yanqi Zhou, Wei Li, and Peter~J Liu.
\newblock Exploring the limits of transfer learning with a unified text-to-text transformer.
\newblock \emph{Journal of machine learning research}, 21\penalty0 (140):\penalty0 1--67, 2020.

\bibitem[Sachdeva et~al.(2024)Sachdeva, Coleman, Kang, Ni, Hong, Chi, Caverlee, McAuley, and Cheng]{how_to}
Noveen Sachdeva, Benjamin Coleman, Wang-Cheng Kang, Jianmo Ni, Lichan Hong, Ed~H Chi, James Caverlee, Julian McAuley, and Derek~Zhiyuan Cheng.
\newblock How to train data-efficient llms.
\newblock \emph{arXiv preprint arXiv:2402.09668}, 2024.

\bibitem[Sakaguchi et~al.(2021)Sakaguchi, Bras, Bhagavatula, and Choi]{winogrande}
Keisuke Sakaguchi, Ronan~Le Bras, Chandra Bhagavatula, and Yejin Choi.
\newblock Winogrande: An adversarial winograd schema challenge at scale.
\newblock \emph{Communications of the ACM}, 64\penalty0 (9):\penalty0 99--106, 2021.

\bibitem[Sap et~al.(2019)Sap, Rashkin, Chen, LeBras, and Choi]{siqa}
Maarten Sap, Hannah Rashkin, Derek Chen, Ronan LeBras, and Yejin Choi.
\newblock Socialiqa: Commonsense reasoning about social interactions, 2019.

\bibitem[Shao et~al.(2024)Shao, Wang, Zhu, Xu, Song, Bi, Zhang, Zhang, Li, Wu, et~al.]{deepseekmath}
Zhihong Shao, Peiyi Wang, Qihao Zhu, Runxin Xu, Junxiao Song, Xiao Bi, Haowei Zhang, Mingchuan Zhang, YK~Li, Y~Wu, et~al.
\newblock Deepseekmath: Pushing the limits of mathematical reasoning in open language models.
\newblock \emph{arXiv preprint arXiv:2402.03300}, 2024.

\bibitem[Shi et~al.(2024)Shi, Zhao, Zhou, and Hao]{industryCorpus2}
Xiaofeng Shi, Lulu Zhao, Hua Zhou, and Donglin Hao.
\newblock Industrycorpus2, 2024.
\newblock URL \url{https://huggingface.co/datasets/BAAI/IndustryCorpus2}.

\bibitem[Shoeybi et~al.(2019)Shoeybi, Patwary, Puri, LeGresley, Casper, and Catanzaro]{megatron}
Mohammad Shoeybi, Mostofa Patwary, Raul Puri, Patrick LeGresley, Jared Casper, and Bryan Catanzaro.
\newblock Megatron-lm: Training multi-billion parameter language models using model parallelism.
\newblock \emph{arXiv preprint arXiv:1909.08053}, 2019.

\bibitem[Soldaini et~al.(2024)Soldaini, Kinney, Bhagia, Schwenk, Atkinson, Authur, Bogin, Chandu, Dumas, Elazar, et~al.]{dolma}
Luca Soldaini, Rodney Kinney, Akshita Bhagia, Dustin Schwenk, David Atkinson, Russell Authur, Ben Bogin, Khyathi Chandu, Jennifer Dumas, Yanai Elazar, et~al.
\newblock Dolma: An open corpus of three trillion tokens for language model pretraining research.
\newblock \emph{arXiv preprint arXiv:2402.00159}, 2024.

\bibitem[Song et~al.(2024)Song, Wang, Zhang, Liu, Lyu, Song, Guo, Yan, Lin, Chen, et~al.]{AlchemistCoder}
Zifan Song, Yudong Wang, Wenwei Zhang, Kuikun Liu, Chengqi Lyu, Demin Song, Qipeng Guo, Hang Yan, Dahua Lin, Kai Chen, et~al.
\newblock Alchemistcoder: Harmonizing and eliciting code capability by hindsight tuning on multi-source data.
\newblock \emph{arXiv preprint arXiv:2405.19265}, 2024.

\bibitem[Suzgun et~al.(2022)Suzgun, Scales, Sch{\"a}rli, Gehrmann, Tay, Chung, Chowdhery, Le, Chi, Zhou, et~al.]{bbh}
Mirac Suzgun, Nathan Scales, Nathanael Sch{\"a}rli, Sebastian Gehrmann, Yi~Tay, Hyung~Won Chung, Aakanksha Chowdhery, Quoc~V Le, Ed~H Chi, Denny Zhou, et~al.
\newblock Challenging big-bench tasks and whether chain-of-thought can solve them.
\newblock \emph{arXiv preprint arXiv:2210.09261}, 2022.

\bibitem[Talmor et~al.(2018)Talmor, Herzig, Lourie, and Berant]{commonsenseqa}
Alon Talmor, Jonathan Herzig, Nicholas Lourie, and Jonathan Berant.
\newblock Commonsenseqa: A question answering challenge targeting commonsense knowledge.
\newblock \emph{arXiv preprint arXiv:1811.00937}, 2018.

\bibitem[Team(2023)]{internlm}
InternLM Team.
\newblock {InternLM}: A multilingual language model with progressively enhanced capabilities, 2023.

\bibitem[Teknium(2023)]{OH25}
Teknium.
\newblock Openhermes 2.5: An open dataset of synthetic data for generalist llm assistants, 2023.
\newblock URL \url{https://huggingface.co/datasets/teknium/OpenHermes-2.5}.

\bibitem[Touvron et~al.(2023)Touvron, Lavril, Izacard, Martinet, Lachaux, Lacroix, Rozi{\`e}re, Goyal, Hambro, Azhar, et~al.]{llama}
Hugo Touvron, Thibaut Lavril, Gautier Izacard, Xavier Martinet, Marie-Anne Lachaux, Timoth{\'e}e Lacroix, Baptiste Rozi{\`e}re, Naman Goyal, Eric Hambro, Faisal Azhar, et~al.
\newblock Llama: Open and efficient foundation language models.
\newblock \emph{arXiv preprint arXiv:2302.13971}, 2023.

\bibitem[Wang et~al.(2024)Wang, Zhang, Wu, Zhao, Shi, Gu, Li, Ma, Pan, and Liu]{cci3}
Liangdong Wang, Bo-Wen Zhang, Chengwei Wu, Hanyu Zhao, Xiaofeng Shi, Shuhao Gu, Jijie Li, Quanyue Ma, TengFei Pan, and Guang Liu.
\newblock Cci3. 0-hq: a large-scale chinese dataset of high quality designed for pre-training large language models.
\newblock \emph{arXiv preprint arXiv:2410.18505}, 2024.

\bibitem[Weber et~al.(2025)Weber, Fu, Anthony, Oren, Adams, Alexandrov, Lyu, Nguyen, Yao, Adams, et~al.]{redpajama}
Maurice Weber, Dan Fu, Quentin Anthony, Yonatan Oren, Shane Adams, Anton Alexandrov, Xiaozhong Lyu, Huu Nguyen, Xiaozhe Yao, Virginia Adams, et~al.
\newblock Redpajama: an open dataset for training large language models.
\newblock \emph{Advances in Neural Information Processing Systems}, 37:\penalty0 116462--116492, 2025.

\bibitem[Wei et~al.(2023)Wei, Zhao, Zhang, Zhu, Wang, Yang, Li, Cheng, L{\"u}, Hu, et~al.]{skypile}
Tianwen Wei, Liang Zhao, Lichang Zhang, Bo~Zhu, Lijie Wang, Haihua Yang, Biye Li, Cheng Cheng, Weiwei L{\"u}, Rui Hu, et~al.
\newblock Skywork: A more open bilingual foundation model.
\newblock \emph{arXiv preprint arXiv:2310.19341}, 2023.

\bibitem[Wenzek et~al.(2019)Wenzek, Lachaux, Conneau, Chaudhary, Guzm{\'a}n, Joulin, and Grave]{ccnet}
Guillaume Wenzek, Marie-Anne Lachaux, Alexis Conneau, Vishrav Chaudhary, Francisco Guzm{\'a}n, Armand Joulin, and Edouard Grave.
\newblock Ccnet: Extracting high quality monolingual datasets from web crawl data.
\newblock \emph{arXiv preprint arXiv:1911.00359}, 2019.

\bibitem[Wettig et~al.(2024)Wettig, Gupta, Malik, and Chen]{qurating}
Alexander Wettig, Aatmik Gupta, Saumya Malik, and Danqi Chen.
\newblock Qurating: Selecting high-quality data for training language models.
\newblock \emph{arXiv preprint arXiv:2402.09739}, 2024.

\bibitem[Xiao et~al.(2024)Xiao, Cai, Zhao, Zeng, Han, Liu, and Sun]{densing_law}
Chaojun Xiao, Jie Cai, Weilin Zhao, Guoyang Zeng, Xu~Han, Zhiyuan Liu, and Maosong Sun.
\newblock Densing law of llms.
\newblock \emph{arXiv preprint arXiv:2412.04315}, 2024.

\bibitem[Yang et~al.(2022)Yang, Hu, Babuschkin, Sidor, Liu, Farhi, Ryder, Pachocki, Chen, and Gao]{mup}
Greg Yang, Edward~J Hu, Igor Babuschkin, Szymon Sidor, Xiaodong Liu, David Farhi, Nick Ryder, Jakub Pachocki, Weizhu Chen, and Jianfeng Gao.
\newblock Tensor programs v: Tuning large neural networks via zero-shot hyperparameter transfer.
\newblock \emph{arXiv preprint arXiv:2203.03466}, 2022.

\bibitem[Yu et~al.(2025)Yu, Dai, Wang, Wang, Chen, and Pei]{chinese_fineweb}
Yijiong Yu, Ziyun Dai, Zekun Wang, Wei Wang, Ran Chen, and Ji~Pei.
\newblock Opencsg chinese corpus: A series of high-quality chinese datasets for llm training.
\newblock \emph{arXiv preprint arXiv:2501.08197}, 2025.

\bibitem[Zellers et~al.(2019)Zellers, Holtzman, Bisk, Farhadi, and Choi]{hellaswag}
Rowan Zellers, Ari Holtzman, Yonatan Bisk, Ali Farhadi, and Yejin Choi.
\newblock Hellaswag: Can a machine really finish your sentence?
\newblock \emph{arXiv preprint arXiv:1905.07830}, 2019.

\bibitem[Zhang et~al.(2024)Zhang, Zeng, Hua, Ding, Chen, Ma, Li, Cui, Qi, Zhu, et~al.]{ultramedical}
Kaiyan Zhang, Sihang Zeng, Ermo Hua, Ning Ding, Zhang-Ren Chen, Zhiyuan Ma, Haoxin Li, Ganqu Cui, Biqing Qi, Xuekai Zhu, et~al.
\newblock Ultramedical: Building specialized generalists in biomedicine.
\newblock \emph{arXiv preprint arXiv:2406.03949}, 2024.

\end{thebibliography}

\clearpage
\appendix
\section{Implementation Details for Efficient Verification}
\label{appendix:implement_details}
All Efficient Verification experiments are trained using the open-source Megatron-LM library. 
We utilize the MiniCPM-1.2B model architecture (as shown in Table 1) with the MiniCPM-3-4B tokenizer. 
The model training utilizes the MiniCPM-3-4B training corpus, and the WSD scheduler. 
We train the model from scratch with 1.1T tokens (1T for the stable stage and 0.1T for the decaying stage). Based on this pretrained model, we further perform a two-stage annealing training with 10B tokens, allocating 30\% of the weight to the verification data, while keeping the remaining 70\% for the default mixed data ratio.
Key training parameters include a sequence length of 4096, weight decay of 0.1, and a gradient clipping threshold of 1.0. We employed a global batch size of 512.
For larger datasets, we train for a total of 5000 steps (approximately 10B tokens). For smaller datasets, we compute the total training steps based on the actual token size of the data and typically allowed the validation data to undergo 3-5 training epochs ($n_{epoch}$).

The training steps calculation formula is as follows:
$$
Total~Iter = \max\left(\frac{Total~Token}{Global~BS \times Seq~Len}, 5000\right)
$$

Where $Total~Token$ is calculated as:
$$
{Total~Token} = \frac{{Curr~Data~Token} \times n_{epoch}}{0.3}
$$

To reduce the cost of baseline experiments, we typically choose training steps of 100, 500, 1,000, 2,500, or 5,000, balancing experimental accuracy and computational resource consumption. 
This means that for datasets of different scales, we dynamically adjust the training steps based on the data tokens required for training. It is important to note that the training steps for the baseline experiments are also dynamically adjusted based on the corresponding dataset size.
Additionally, we set the warmup fraction to 0.1, and the annealing phase used an exponential decay approach, with the maximum learning rate to 1e-3 and the minimum learning rate to 5e-5. To enhance training stability, we use Maximal Update Parameterization (MuP).

\section{Results and Analysis of Efficient Verification}
\label{appendix:efficient_verification_result}
To verify the effectiveness of the efficient verification strategy, we use the FineWeb and FineWeb-edu datasets, training both on the efficient verification and from-scratch 100B token strategies, and compare the results.
For evaluation, we use OpenCompass~\citep{opencompass} for the model trained with the efficient verification strategy, and Lighteval~\citep{lighteval} for the model trained from scratch with 100B tokens. The experimental results are shown in Table~\ref{tab:appendix_efficient_verification_result}. 

\begin{table}[!htb]
    \centering
    \small
    \captionsetup{justification=centerlast, singlelinecheck=false}
    \caption{Comparison of efficient verification strategy and from-scratch 100B token strategies on FineWeb and FineWeb-edu.}
    \renewcommand{\arraystretch}{1.2}
    \begin{tabular}{lcccccc}
        \toprule
        & \multicolumn{3}{c}{Efficient Verification} & \multicolumn{3}{c}{100B From Scratch} \\
        \cmidrule(lr){2-4} \cmidrule(lr){5-7}
        Metrics  & FineWeb & FineWeb-edu & Diff. & FineWeb  & FineWeb-edu & Diff. \\
        \midrule
        MMLU         & 45.84 & 47.35 & \textcolor{red}{+1.51}       & 28.84 & 31.80 & \textcolor{red}{+2.96}    \\
        HellaSwag    & 57.72 & 56.99 & \textcolor{blue}{-0.73}      & 42.91 & 42.17 & \textcolor{blue}{-0.74}   \\
        ARC-C        & 38.98 & 39.66 & \textcolor{red}{+0.68}       & 25.17 & 34.56 & \textcolor{red}{+9.39}    \\
        ARC-E        & 57.67 & 59.08 & \textcolor{red}{+1.41}       & 59.18 & 69.95 & \textcolor{red}{+10.77}   \\
        PIQA         & 74.48 & 72.91 & \textcolor{blue}{-1.57}      & 73.29 & 72.14 & \textcolor{blue}{-1.15}   \\
        SIQA         & 43.55 & 43.35 & \textcolor{blue}{-0.20}      & 38.95 & 38.13 & \textcolor{blue}{-0.82}   \\
        Winogrande   & 56.67 & 55.56 & \textcolor{blue}{-1.11}      & 55.64 & 55.56 & \textcolor{blue}{-0.08}   \\
        OpenbookQA   & 66.80 & 69.40 & \textcolor{red}{+2.60}       & 22.20 & 25.20 & \textcolor{red}{+3.00}    \\
        \midrule
        \textit{Average}      & 55.21 & 55.54 & \textcolor{red}{+0.33}       & 41.00 & 43.00 & \textcolor{red}{+2.00}    \\
        \bottomrule
    \end{tabular}
    \label{tab:appendix_efficient_verification_result}
\end{table}

We can observe that the efficient verification strategy exhibited consistent trends across multiple evaluation tasks when compared to the from-scratch 100B model. 
For example, in metrics like MMLU, ARC-E, ARC-C, and OpenbookQA, FineWeb-edu consistently outperformed FineWeb under both training paradigms. 
Similarly, for metrics such as HellaSwag, PIQA, SIQA, and Winogrande,  FineWeb-edu showed performance degradation compared to FineWeb, regardless of training strategy.
Overall, the efficient verification strategy quickly revealed the impact of the validation data on various evaluation dimensions and provided accurate feedback. 
This strategy significantly reduces computational resource requirements, enabling more efficient data quality assessment and optimization, ultimately enhancing model training effectiveness.

\end{document}